\documentclass[preprint,review,10pt]{elsarticle}

\usepackage{amssymb}
\usepackage{amsmath}
\usepackage{amsthm}

\usepackage{lineno}

\biboptions{numbers,sort&compress}
\usepackage{subfigure}
\usepackage{booktabs}
\usepackage{multirow}
\usepackage{color}

\def\etal{{\emph{et al. }}}

\makeatletter
\renewcommand*\env@matrix[1][\arraystretch]{%
\edef\arraystretch{#1}%
\hskip -\arraycolsep
\let\@ifnextchar\new@ifnextchar
\array{*\c@MaxMatrixCols c}}
\makeatother

\journal{}

\begin{document}

\begin{frontmatter}



\title{Two-stage Rule-induction Visual Reasoning on RPMs with an Application to Video Prediction}


\author[add1]{Wentao He}
\author[add1]{Jianfeng Ren\corref{cor}}
\ead{jianfeng.ren@nottingham.edu.cn}
\author[add1]{Ruibin Bai}
\author[add2]{Xudong Jiang}
\cortext[cor]{Corresponding author. Tel.: +86 (0)574 8818 0000--8805}

\address[add1]{The School of Computer Science, University of Nottingham Ningbo China, 199 Taikang East Road, Ningbo, 315000 China}
            
\address[add2]{School of Electrical \& Electronic Engineering, Nanyang Technological University, Nanyang Link, 639798 Singapore}

\begin{abstract}
Raven's Progressive Matrices (RPMs) are frequently used in evaluating human's visual reasoning ability. Researchers have made considerable efforts in developing systems to automatically solve the RPM problem, often through a black-box end-to-end convolutional neural network for both visual recognition and logical reasoning tasks. Based on the two intrinsic natures of RPM problem, visual recognition and logical reasoning, we propose a Two-stage Rule-Induction Visual Reasoner (TRIVR), which consists of a perception module and a reasoning module, to tackle the challenges of real-world visual recognition and subsequent logical reasoning tasks, respectively. For the reasoning module, we further propose a ``2+1'' formulation that models human's thinking in solving RPMs and significantly reduces the model complexity. It derives a reasoning rule from each RPM sample, which is not feasible for existing methods. As a result, the proposed reasoning module is capable of yielding a set of reasoning rules modeling human in solving the RPM problems. To validate the proposed method on real-world applications, an RPM-like Video Prediction (RVP) dataset is constructed, where visual reasoning is conducted on RPMs constructed using real-world video frames. Experimental results on various RPM-like datasets demonstrate that the proposed TRIVR achieves a significant and consistent performance gain compared with the state-of-the-art models. 
\end{abstract}




\end{frontmatter}


\section{Introduction}
\label{sec1}
Computer vision has recently achieved great advancements in many applications \cite{he2016deep,ren2015learning,ren2015faster,ren2013complete}. The research focus has gradually shifted from visual recognition of individual objects to visual understanding of scene images/videos. Visual reasoning \cite{zellers2019recognition} is one of the visual understanding tasks, which usually consists of two parts: ``visual'' and ``reasoning''. The former relates to a visual system to gain information through vision ability \cite{he2016deep,ren2015faster,redmon2016you}, and the latter links to a cognitive system, focusing on using logic to discover rules and solve problems \cite{santoro2017simple,messina2018learning,zhang2019raven,santoro2018measuring,sekh2020can,palm2018recurrent,battaglia2016interaction}. 
Significant research efforts are needed to develop a visual reasoning system that has not only the visual recognition capability, but also the capability of understanding image/video scenes and conducting logical reasoning based on the perceived image/video contents. 

Visual reasoning in general can be categorized as: video action recognition \cite{wang2016temporal,zhou2018temporal,lin2019tsm}, image/video captioning \cite{wang2020learning,yao2018exploring}, visual question answering (VQA) \cite{antol2015vqa,johnson2017clevr,wang2017fvqa,zhu2016visual7w}, and visual IQ tests \cite{santoro2018measuring,zhang2019raven,sekh2020can}. The first two focus more on visual recognition whereas the last two focus more on logical reasoning. Early models for action recognition and VQA develop reasoning abilities by building task-specific multilayer perceptrons or fine-tuning visual recognition models \cite{johnson2017inferring}. With the development of Relation Networks \cite{santoro2017simple} and Graph Neural Networks \cite{garcia2017few}, models with better reasoning abilities have been developed, but mostly for simple reasoning logic \cite{wang2018deep,garcia2017few}.

\begin{figure}[htbp]
	\centering
	\subfigure[Question]{
		\begin{minipage}[t]{0.38\linewidth}
			\centering
			\includegraphics[width=3.3cm]{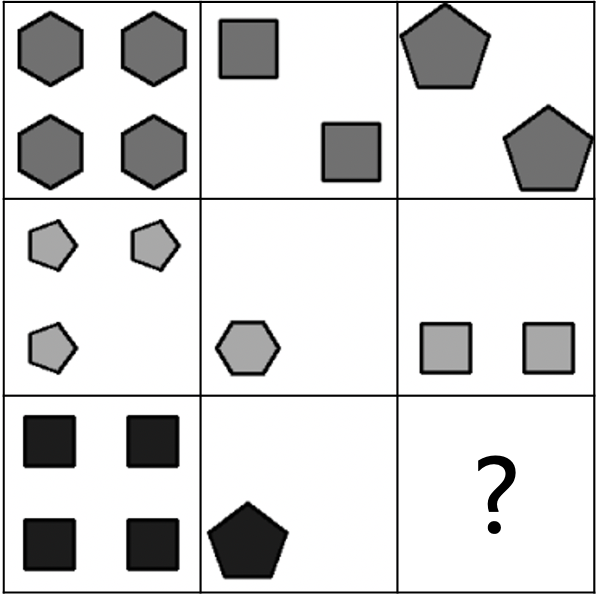}
		\end{minipage}
	}
	\subfigure[Candidate answers]{
		\begin{minipage}[t]{0.53\linewidth}
			\centering
			\includegraphics[width=4.4cm]{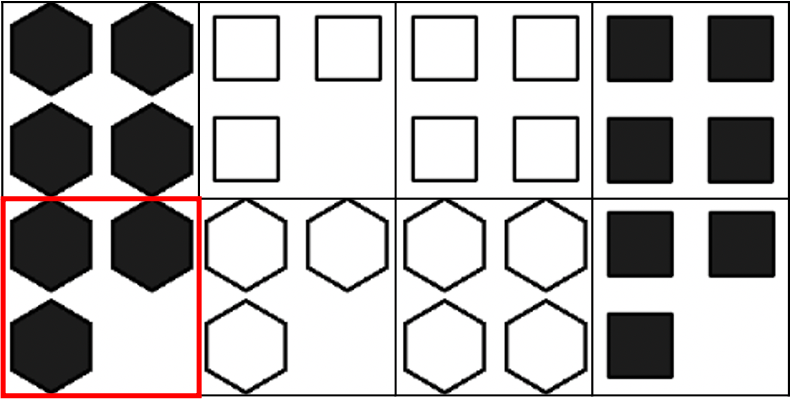}
		\end{minipage}
	}
	\centering
	\caption{An example of the I-RAVEN dataset \cite{hu2021stratified}. The correct answer is marked in red. The reasoning rules are derived row-wisely based on the extracted visual attributes, i.e. \textit{Arithmetic} (subtraction) on \textit{Number}, \textit{Distribute\_Three} on \textit{Type}, \textit{Progression} on \textit{Size} and \textit{Constant} on \textit{Color}. }
	\label{example}
\end{figure}

Logical reasoning consists of three taxonomies of reasoning, i.e., induction, deduction and abduction \cite{reichertz2013induction}. Generic logical reasoning contains a premise (precondition), a conclusion (logical consequence) and a rule (material conditional) that implies the conclusion given the premise.
Reasoning on complex image scenes is tremendously difficult \cite{zellers2019recognition}, and hence very often the visual task is either simplified, e.g. Sort-of-CLEVR \cite{santoro2017simple,messina2018learning} and Pretty-CLEVR \cite{palm2018recurrent}, or the visual information is assumed available in advance \cite{santoro2017simple,palm2018recurrent,battaglia2016interaction}. One test frequently used on human's visual reasoning ability in cognitive science is Raven's Progressive Matrices (RPM) \cite{mcgreggor2014fractals}. Researches into solving RPM automatically can help to assess and improve the reasoning ability of computer vision systems. In literature, RPMs are often referred as abstract reasoning \cite{hu2021stratified,santoro2018measuring,Wang2020Abstract} or analogical reasoning \cite{mcgreggor2014fractals,kim2020few,zhang2019raven}, and both of which enable a form of inductive reasoning \cite{reichertz2013induction}.
An RPM problem is usually formed by a $3\times 3$ pictorial matrix with the last one left blank. The task is to infer the underlying rules from the given matrix and identify the missing entry from the list of given candidates. Fig.~\ref{example} shows an example of RPM problems. 

Researchers have made considerable efforts in developing a system to automatically solve the RPM problem \cite{zhang2019learning,zheng2019abstract,kim2020few,Wang2020Abstract}. Existing approaches often offer a black-box solution, which is difficult to be comprehended by humans and extendable to real-world problems. 1) Most existing solutions utilize end-to-end convolutional neural networks (CNNs) \cite{zhang2019learning,hu2021stratified,zheng2019abstract,kim2020few,Wang2020Abstract,an2020hierarchical}. Such a network may not be able to handle the tremendous combinations of complex visual recognition tasks and diversified logical reasoning tasks. 2) Most existing methods \cite{zhang2019learning,hu2021stratified,zheng2019abstract,kim2020few,Wang2020Abstract,an2020hierarchical} aim to discover the underlying reasoning rule that directly maps from eight problem images to one of the candidate answers. Such ``8+1'' formulation neglects the fact that humans often solve RPM-like problems row-wisely or column-wisely. As shown later in Section~\ref{sec3.c}, this formulation greatly enlarges the input to the system and unnecessarily complicates the model. 3) Due to the ``8+1'' formulation, it is difficult or even impossible to extract any precise logical reasoning rules in RPM-like problems. 
Those rules are overwhelmed in the existing black-box solutions \cite{zhang2019learning,hu2021stratified,zheng2019abstract,kim2020few,Wang2020Abstract,an2020hierarchical}. 

To address these challenges, we propose a solution simulating human's visual reasoning process, named as Two-stage Rule-Induction Visual Reasoner (TRIVR). 1) To handle complex and diversified visual reasoning tasks, instead of an end-to-end network, we propose a two-stage visual reasoning framework. The first stage, a perception module, aims to visually understand the image scenes and extract visual attributes. The second stage, a reasoning module, aims to build reasoning rules based on the derived visual attributes. This two-stage design could make good use of existing developments in both computer vision and logical reasoning societies, and well handle the complicated and diversified real-world visual reasoning tasks. 2) In contrast to the existing ``8+1'' formulation \cite{zhang2019learning,hu2021stratified,zheng2019abstract,kim2020few,Wang2020Abstract,an2020hierarchical}, we formulate the problem as three groups of ``2+1'' image sets for each problem panel row-wisely, i.e. based on the first two images, the target is to derive the underlying rules that define the third image, and the rules should be consistent across three groups. This is very close to human's reasoning process. As illustrated later in Section~\ref{sec3.c}, one reasoning rule could be derived from each RPM sample by utilizing the proposed formulation, which is not feasible for the existing ``8+1'' formulation. 3) Beneficial from the proposed ``2+1'' formulation, the proposed approach could produce a set of reasoning rules that that are consistent with human's reasoning process, 
as illustrated later in Section~\ref{sec:3.d}. 


RPM problems are challenging, 
e.g., given the first two rows (each with three images) only, a set of consistent rules need to be discovered from images and applied to the third row to find the last missing image. Although RPM problems are artificially generated, they have been used to evaluate the reasoning ability of humans and various visual reasoning models \cite{zhang2019learning,hu2021stratified,zheng2019abstract,kim2020few,Wang2020Abstract,an2020hierarchical}. To evaluate the proposed method in real-world applications, we construct an RPM-like Video Prediction (RVP) dataset. 
The video frames of the UA-DETRAC dataset \cite{wen2020ua}, a benchmark real-world multi-object detection and tracking dataset, are used to construct the RPM problems. The target here is to predict the future frame given the two previous frames at intervals of time, together with the first two frame triplets from other videos. To simplify the task, eight candidate answers are provided. The problem is formulated under the RPM framework to assess the visual reasoning capability of the proposed method and many other models. More details can be found in Section~\ref{sec4}.

The proposed approach is compared with state-of-the-art methods 
on the two widely used benchmark datasets, RAVEN \cite{zhang2019raven} and I-RAVEN \cite{hu2021stratified}, their shrunk versions and the developed RVP dataset to evaluate the reasoning capability of models. The proposed method significantly improves the average visual reasoning accuracy from 91.4\% to 93.1\% compared with the previously published best results by CoPINet \cite{zhang2019learning} on the RAVEN dataset, and from 60.8\% to 95.9\% compared with the previously published best result by SRAN \cite{hu2021stratified} on the I-RAVEN dataset. On the RVP dataset, the proposed method outperforms CoPINet \cite{zhang2019learning} by 9.24\% and SRAN \cite{hu2021stratified} by 7.69\%.

Our contributions are summarized as follow: 1) We propose a two-stage framework making good use of developments in both computer vision and logical reasoning societies and solving the RPM problem by simulating the human's solving process. 2) The proposed ``2+1'' formulation generates one reasoning rule by utilizing each sample question, which greatly simplifies the model and yields an elegant and efficient solution. 3) Thanks to the two-stage framework and ``2+1'' formulation, the proposed solution could yield a set of reasoning rules adaptive to various scenarios covering almost all the reasoning rules in RPM-like problems. 4) We develop a benchmark RPM-like Video Prediction dataset and successfully apply the proposed method to solve the complicated real-world visual reasoning problems.

\section{Related Works}
\label{sec2}
\subsection{Visual Reasoning}
Visual understanding has been studied for decades, while visual reasoning only recently attracted the attention of researchers. 
It first visually recognizes the attributes from image/video scenes \cite{ren2013complete,ren2015learning} and then conducts relational reasoning based on the derived attributes \cite{santoro2017simple,messina2018learning,zhang2019raven,santoro2018measuring,palm2018recurrent,battaglia2016interaction}. Visual reasoning requires efforts from both computer science and cognitive science, and develops linkages between recognition and cognition \cite{zellers2019recognition}. 
In literature, visual reasoning spans a variety of tasks, e.g., action recognition \cite{wang2016temporal,zhou2018temporal,lin2019tsm}, image/video captioning \cite{wang2020learning,yao2018exploring}, visual question answering \cite{antol2015vqa,johnson2017clevr,wang2017fvqa,zhu2016visual7w,yu2020cross,liu2021dual}, and Raven's Progressive Matrices \cite{mcgreggor2014fractals,zhang2019learning,hu2021stratified,zheng2019abstract,kim2020few,Wang2020Abstract,an2020hierarchical}. 

Activity recognition highly relies on temporal information. To efficiently recognize actions in video sequences, reasoning about human-object relations over time becomes a crucial challenge \cite{zhou2018temporal}. Wang \etal \cite{wang2016temporal} introduced Temporal Segment Network based on BN-Inception architecture. Zhou \etal \cite{zhou2018temporal} improved it by utilizing Temporal Relation Network for relational reasoning and learning temporal dependencies between video frames at multiple time scales. 

Image/video captioning from a relation-reasoning perspective has recently received an increasing attention \cite{hou2020joint}. Yao \etal \cite{yao2018exploring} developed an image captioning scheme based on semantic and spatial object relationships, with a relation-aware region representation derived by Graph Convolutional Network, which generates captions by utilizing the region-level attention mechanism of LSTM. Hou \etal \cite{hou2020joint} built models exploiting relationships between scene objects and prior knowledge for image and video captioning.

Visual question answering (VQA) is a conventional visual reasoning task that measures the machine understanding of scene-level images. The objective is to derive an accurate natural language answer, given an image and a related natural language question \cite{antol2015vqa}. Early VQA tasks are based on natural scene images \cite{antol2015vqa,zhu2016visual7w,goyal2017making}. Johnson \etal \cite{johnson2017clevr} developed the CLEVR dataset by replacing natural images with synthetic images to address the challenge of complicated background in natural images. Wang \etal \cite{wang2017fvqa} developed the fact-based VQA that contains supportive natural language facts to assist reasoning. Recently, a new form of VQA tasks was developed by Zellers \etal as Visual Commonsense Reasoning \cite{zellers2019recognition}, whose objective is to answer the question and/or provide an explanation why the answer is correct. 

\subsection{Raven's Progressive Matrices}
Raven's Progressive Matrices are originally designed as a non-verbal assessment for human intelligence. By using pictorial matrices containing visually simple patterns, it minimizes the impact of language barrier and culture bias. Recently, large-scale RPM-style datasets, RAVEN \cite{zhang2019raven} and its variants \cite{hu2021stratified}, were introduced to extend the RPM study from cognitive science to computer science. These datasets \cite{zhang2019raven,hu2021stratified} are often automatically generated, and the problems are solved with minimal prior knowledge about the internal construction rules. 

Santoro \etal~\cite{santoro2018measuring} developed a Wild Relation Network (WReN), which applies a Relation Network to model the relationship between the question panel and the candidate answer. Zhang \etal~\cite{zhang2019raven} developed a ResNet architecture assembled with a Dynamic Residual Tree (DRT) module for reasoning \cite{zhang2019raven}. They later developed an improved network architecture called CoPINet~\cite{zhang2019learning} under the principle of contrasting, which achieved the previous best performance on the RAVEN dataset. 
Hu \etal \cite{hu2021stratified} argued that RAVEN has a shortcut to solutions, and hence developed an I-RAVEN dataset balancing the candidate answer sets and the solution model. Their developed solution model, SRAN \cite{hu2021stratified}, achieves the state-of-the-art results on the I-RAVEN dataset.

Various models have been developed in literature to solve the RPM-like problems \cite{hu2021stratified,zheng2019abstract,kim2020few,Wang2020Abstract,an2020hierarchical}. However, most of them are based on the black-box CNN models not understandable by humans, whereas the RPM-like problems are originally designed to evaluate human intelligence and mostly solvable by humans. In this paper, we propose a framework simulating human's reasoning process, which outperforms all the existing solution models.

\subsection{Video Prediction}
Video prediction is an emerging research field of computer vision~\cite{oprea2020review}. It has been successfully applied in various applications such as action anticipation \cite{gammulle2019predicting}, prediction of object locations \cite{makansi2019overcoming}, trajectory prediction \cite{bhattacharyya2018long} and anomaly detection \cite{liu2018future}. Given a sequence of previous frames, the target of video prediction is to reason and predict about the subsequent frame(s) based on the analysis of rich spatio-temporal features in a video, e.g., object/background information or regularity of pixel changes \cite{oprea2020review}.

In literature, many solution models 
have been designed to tackle the video prediction problem, which can be broadly classified into four categories \cite{oprea2020review}: direct pixel synthesis \cite{jin2020exploring}, factoring the prediction space \cite{chen2019uni}, narrowing the prediction space \cite{villegas2017learning} and incorporating uncertainty \cite{goroshin2015learning}. For instance, Jin \etal \cite{jin2020exploring} developed a generative adversarial network to synthesize future frames based on multi-level wavelet analysis, given a set of previous frames. To reduce the prediction space, Villegas \etal \cite{villegas2017learning} extracted the human pose as an additional supervision signal, and regressed future frames through analogy making. Recently, Chen \etal \cite{chen2019uni} predicted motions of different key objects to generate sharper and more realistic frame predictions. For a more comprehensive review on video prediction, readers are referred to \cite{oprea2020review}.

\section{Reasoning on Raven's Progressive Matrices}
\label{sec3}
\subsection{Proposed Two-step Framework}
Formally, in a $3\times3$ RPM-like problem, the query image matrix $\boldsymbol{I}^q$ and the candidate answer set $\boldsymbol{I}^c$ can be represented as: 
\begin{align}
	& \boldsymbol{I}^q=\begin{bmatrix}[1.5]
		{I}^q_{1,1} & {I}^q_{1,2} & {I}^q_{1,3} \\
		{I}^q_{2,1} & {I}^q_{2,2} & {I}^q_{2,3} \\
		{I}^q_{3,1} & {I}^q_{3,2} & {I}^q_{3,3} \\
	\end{bmatrix},  \\
	& \boldsymbol{I}^c=\{{I}^c_1, {I}^c_2, \dots, {I}^c_8\}, 
\end{align}
where the last missing image ${I}^q_{3,3}$ is filled by the ground-truth answer ${I}^c_*$ from the candidate answer set $\boldsymbol{I}^c$, as shown in Fig.~\ref{example}. 
A set of questions with known correct candidate answers serve as training samples $\langle\boldsymbol{I}^q_{(i)},\boldsymbol{I}^c_{(i)}\rangle, i=1,\dots,n$, where $n$ is the size of the training set. The objective is to identify the correct candidate image ${I}^c_*$ from $\boldsymbol{I}^c$ based on the derived attribute matrix $\boldsymbol{A}^q$ and the candidate attribute vector $\boldsymbol{a}^c$, together with the underlying reasoning rule ${f}$: 
\begin{align}\label{a_matrix}
	& \boldsymbol{A}^q = \begin{bmatrix}[1.5]
		a^q_{1,1} & a^q_{1,2} & a^q_{1,3} \\
		a^q_{2,1} & a^q_{2,2} & a^q_{2,3} \\
		a^q_{3,1} & a^q_{3,2} & a^q_{3,3} \\
	\end{bmatrix},  \\
	& \boldsymbol{a}^c = \{a^c_1, a^c_2, \dots, a^c_8\},
\end{align}
where visual attributes $a^q, a^c \in \{N, P, T, S, C\}$ are either the attributes of \textit{Number}, \textit{Position}, \textit{Type}, \textit{Size} or \textit{Color} of a question/candidate image, which will be derived through the perception module.

For humans, the problems are usually solved in two steps: the visual attributes are first derived from input images through the visual perception system, and then reasoning rules are derived based on the extracted visual attributes through the logical reasoning system. When solving RPMs, humans may try a series of logical reasoning rules generated by human's cognition system, and apply the most appropriate one to predict the correct answer. 

For most existing methods, visual recognition and logic reasoning are assembled into one CNN architecture \cite{zhang2019learning,hu2021stratified,zheng2019abstract,kim2020few,Wang2020Abstract,an2020hierarchical}. Such techniques may sound technically appealing and may work in a situation where either visual recognition or logical reasoning is relatively easy. However, often in visual reasoning, the image scenes are so complicated \cite{wang2016temporal,zhou2018temporal,lin2019tsm,yu2020cross,wang2020learning,yao2018exploring} that a dedicated system is needed to accomplish a specific visual recognition task, and many different systems of such types are needed for various tasks. In addition, the reasoning modules would be different for different reasoning tasks. Therefore, it is extremely difficult to build an end-to-end architecture that could work well for various visual reasoning tasks. 

\begin{figure*}[ht]
	\centering
	\includegraphics[width=1\textwidth]{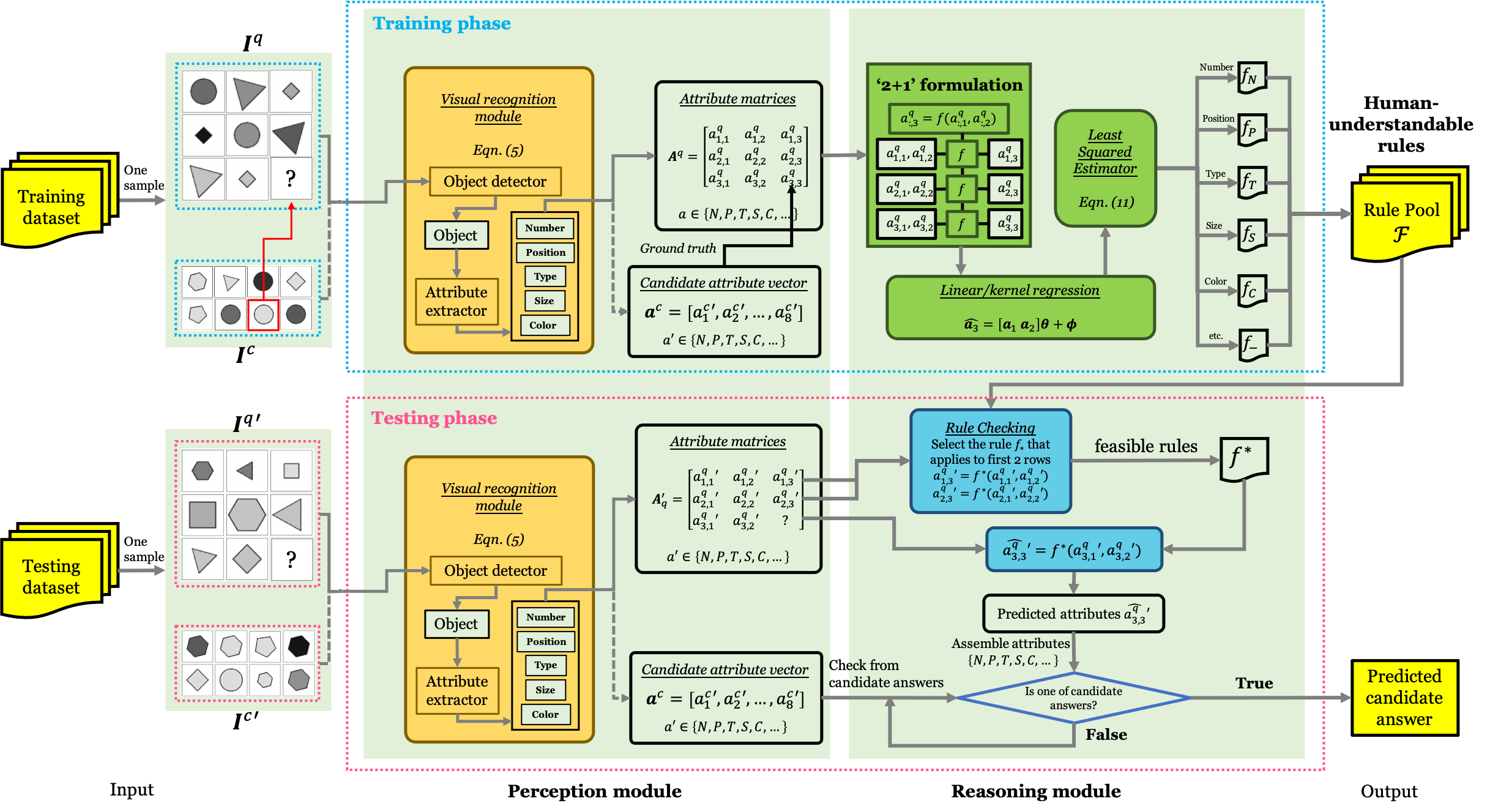}
	\caption{Block diagram of the proposed TRIVR. The visual attributes are first derived by the perception module, and used to extract the reasoning rules by the reasoning module. By utilizing the proposed ``2+1'' formulation, one reasoning rule can be extracted by a least-squares estimator as in Eqn. (\ref{eq:ls}) for each attribute using one sample problem only. During testing, the visual attributes of a testing sample are first derived by the perception module and then evaluated against the rules in the rule pool $\mathcal{F}$ using Eqn. (\ref{eq:1}). The proposed method yields an efficient and elegant solution model for solving RPM problems. }
	\label{block-diagram}
\end{figure*}

In contrast, the proposed TRIVR solves the visual reasoning problems by utilizing two separate modules, the perception module and the reasoning module, as shown in Fig. \ref{block-diagram} The former aims to address the challenges of deriving the visual attributes from the image/video, and the latter aims to tackle the challenges of deriving the logical reasoning rules based on the perceived visual attributes. By disassembling the whole framework into two modules, we could make good use of the recent developments in both visual recognition and logic reasoning to better solve the visual reasoning tasks.

\noindent {\textbf{Training phase}} 
The proposed framework consists of two stages: perception and reasoning. The objective of the first stage is to develop the perception module $M_P$ that transfers the input images $\boldsymbol{I}^q$ and $\boldsymbol{I}^c$ into the set of visual attributes $\boldsymbol{A}^q$ and $\boldsymbol{a}^c$ later to be fed into the reasoning module $M_R$. More information on the perception module can be found in Section \ref{sec.3.b}. 
The objective of the second stage is to develop the reasoning module $M_R$ that infers a set of reasoning rules based on the derived visual attributes $\boldsymbol{A}^q$ and $\boldsymbol{a}^c$ using the proposed ``2+1'' formulation. More specifically, for each row, given the visual attributes of the first two images, the target is to find the underlying rule $f$ to derive the attributes of the third image. Compared to the ``8+1'' formulation used in existing methods, the proposed ``2+1'' formulation greatly simplifies the model and leads to an elegant solution model, i.e. one reasoning rule could be derived for each attribute using one sample question only. These reasoning rules then form the pool of reasoning rules $\mathcal{F}$, to be later used in testing.
More details on logical reasoning can be found in Section~\ref{sec3.c}.

\noindent {\textbf{Testing phase}} 
During testing, given an image matrix ${\boldsymbol{I}^q}^{\prime}$ and candidate image set ${\boldsymbol{I}^c}^\prime$, their corresponding visual attributes ${\boldsymbol{A}^q}^\prime$ and ${\boldsymbol{a}^c}^\prime$ are first derived using the proposed perception module $M_P$. At the reasoning stage, following the proposed ``2+1'' formulation, the target is to find a consistent rule $f^{*}$ from the rule pool $\mathcal{F}$ to apply on the first two rows. Note that more than one feasible rule $f^*$ may be found at this point. Then all the feasible rules are applied on the third row to predict the correct candidate answer. If the predicted answer matches any candidate answer, it is treated as the final answer.

The proposed two-step framework for visual reasoning is consistent to human's visual reasoning. It addresses the challenges of poor interpretablity of existing end-to-end networks for visual reasoning. Moreover, complicated visual reasoning tasks involving complex visual recognition and diversified logical reasoning rules are potentially better solved in two steps rather than using one end-to-end network, as indicated later in the experiments. 

\subsection{Perception Module}\label{sec.3.b}
The aim of the perception module $M_P$ is to extract the visual attributes $a^q,a^c$ from the question image $I^q$ or the candidate image $I^c$, i.e. 
\begin{align}
	{a}^q = M_P({I}^q), \nonumber\\
	{a}^c = M_P({I}^c).
\end{align}

In literature, many existing solutions can be used to derive these attributes \cite{he2016deep,ren2015faster,redmon2016you}. In this paper, computationally efficient methods are utilized to extract the visual attributes from the images. 
The \textit{Type} and \textit{Size} attributes are determined using template matching. The \textit{Number} and \textit{Position} attributes are derived through Connected Component Analysis, and the \textit{Color} attribute are derived as the median intensity value of the polygon.

The derived perception module is utilized in both training and test phases to derive the visual attributes $\boldsymbol{A}^q$ for the input image matrix and $\boldsymbol{a}^c$ for the candidate images. The derived attributes have different meanings and should be handled differently with care. Attribute \textit{Number}, \textit{Type}, \textit{Size} and \textit{Color} are formed by a set of ordered values. For attribute \textit{Position}, its values indicate whether there is an object at the corresponding {position}, e.g. for one $2\times2$ image sub-patch with \textit{Position} vector [{1},{0},{0},{1}], {1} indicates that there is an object and {0} otherwise. For all visual attributes, the categorical attribute values are transformed into numerical values. 

The developed perception module perfectly recognizes all the attributes, \textit{Type}, \textit{Number/Position}, \textit{Size}, \textit{Color}, with an accuracy of 100\%. The derived visual attributes are then fed into the reasoning module to discover the underlying rules governing the sample question.

\subsection{Reasoning Module using Proposed ``2+1'' Formulation}
\label{sec3.c}

Traditionally, human solves RPM problems by finding a common rule $R_H$ that simultaneously applies to all three rows,
\begin{align}
	{a^q_{i,3}} = {R_H}\left(a^q_{i,1},a^q_{i,2}\right),
\end{align}
where $a^q_{i,1}$, $a^q_{i,2}$, $a^q_{i,3}$ represent 3 attribute values of images in the $i^\text{th}$ row. 
Following human's reasoning process, we propose the ``2+1'' formulation. We redefine the $3 \times 3$ attribute matrix $\boldsymbol{A}^q$ as follows: 
\begin{align}\label{concat}
	& \boldsymbol{A}^q=\begin{bmatrix}
		\boldsymbol{a}^q_{1} & \boldsymbol{a}^q_2 & \boldsymbol{a}^q_3
	\end{bmatrix}, \nonumber \\
	& \boldsymbol{A}^q_{1,2} = \begin{bmatrix}
		\boldsymbol{a}^q_{1} & \boldsymbol{a}^q_2 
	\end{bmatrix}, \nonumber \\
	& \text{where } \boldsymbol{a}^q_{1}=\begin{bmatrix}[1.5]
		a^q_{1,1}\\
		a^q_{2,1}\\
		a^q_{3,1}
	\end{bmatrix}, \boldsymbol{a}^q_{2}=\begin{bmatrix}[1.5]
		a^q_{1,2} \\
		a^q_{2,2} \\
		a^q_{3,2}
	\end{bmatrix} \text{ and } \boldsymbol{a}^q_{3}=\begin{bmatrix}[1.5]
		a^q_{1,3} \\
		a^q_{2,3} \\
		a^q_{3,3}
	\end{bmatrix}.
\end{align}

The RPM problems is transferred into a small-scale regression problem, i.e. for each question panel,  given the input matrix $\boldsymbol{A}^q_{1,2}$, the objective is to derive a regression rule $f(\cdot)$ so that $\hat{\boldsymbol{a}}^q_{3} = f(\boldsymbol{A}^q_{1,2})$, and the difference between $\hat{\boldsymbol{a}}^q_{3}$ and $\boldsymbol{a}^q_{3}$ should be minimized. 
Our hypothesis starts with linear regression, as the reasoning rules defined in \cite{zhang2019raven} are mainly unary operations, i.e. 
\begin{align}\label{eq:1}
	\hat{\boldsymbol{a}}^q_{3} = f\left(\boldsymbol{A}^q_{1,2}\right) = \boldsymbol{A}^q_{1,2} \boldsymbol{\theta} + \boldsymbol{\phi},
\end{align}
where $\boldsymbol{\theta} \in \mathbb{R}^2$ is a vector of linear coefficients and $\boldsymbol{\phi} \in \mathbb{R}^3$ is the offset. The set of coefficients $\boldsymbol{\theta}$ and $\boldsymbol{\phi}$ that satisfy the rule $f(\cdot)$ can be derived by solving the following optimization function: 
\begin{align}\label{eq:4}
	\min_{\boldsymbol{\theta}, \boldsymbol{\phi}} \left|\left|\boldsymbol{a}^q_{3}-( {\boldsymbol{A}^q_{1,2}} \boldsymbol{\theta} + \boldsymbol{\phi})\right|\right|.
\end{align}
We could solve the optimization problem by least-squares estimator (LSE), i.e.
\begin{align}\label{eq:ls}
	{\boldsymbol{\theta}}_{\text{LS}} &= \left({\boldsymbol{A}^q_{1,2}}^\top \boldsymbol{A}^q_{1,2}\right)^{-1} {\boldsymbol{A}^q_{1,2}}^\top \boldsymbol{a}^q_3, \nonumber \\
	{\boldsymbol{\phi}} &= \boldsymbol{a}^q_3 - \boldsymbol{A}^q_{1,2} {\boldsymbol{\theta}}_{\text{LS}}. 
\end{align}

To get the closed-form solution for ${\boldsymbol{\theta}}_{\text{LS}}$, ${\boldsymbol{A}^q_{1,2}}^\top \boldsymbol{A}^q_{1,2}$ should have full rank to derive the matrix inverse. The rank of \textit{Constant} matrices is always 1, while \textit{Progression}, \textit{Arithmetic} and \textit{Distribute\_Three} matrices have full rank. Thus, the pseudo-inverse is used for \textit{Constant} matrices. We will further provide a mathematical inference of conditions that need to be satisfied for each configuration respectively in the next subsection.

The proposed ``2+1'' formulation simulates human's reasoning process, and it is more efficient and elegant to solve. In contrast, the ``8+1'' formulation learns a rule $g_{\boldsymbol{w}}(\cdot)$ by CNN architectures with weights $\boldsymbol{w}$ that fulfill: 
\begin{align}\label{eq:2}
	{a}^q_{3,3} = g_{\boldsymbol{w}}\left(a^q_{1,1},a^q_{1,2},a^q_{1,3},a^q_{2,1},a^q_{2,2},a^q_{2,3},a^q_{3,1},a^q_{3,2}\right).
\end{align}
Compared to the proposed ``2+1'' formulation, it unnecessarily complicates the RPM problems by using a sophisticated formulation and solves it using a much more complicated CNN architecture. One exact reasoning rule can be extracted based on one sample question using the proposed ``2+1'' formulation, whereas the ``8+1'' formulation needs much more samples to derive the rule. In addition, the extracted rules from the ``2+1'' formulation can be easily interpreted by humans, whereas the ``8+1'' can not. What's more, we could reason the correct answer without the help of negative candidate answers using the ``2+1'' formulation, whereas it is very difficult or even infeasible for the ``8+1'' formulation.

\subsection{Inference of Human-understandable Rules}
\label{sec:3.d}

In this section, we show that the rules inferred from Eqn. (\ref{eq:ls}) can be transferred to a set of human-understandable rules, and the rules used in the RAVEN problems such as \textit{Constant}, \textit{Progression}, \textit{Arithmetic} and \textit{Distribute\_Three} can be well represented by the rules derived from Eqn. (\ref{eq:ls}). 
As a result, humans could have more insights of the underlying reasoning process and at the time have more confidence on the reasoning outcomes of the model. Take note that when the rules are extracted using Eqn. (\ref{eq:ls}), we do not require any additional prior knowledge on the underlying rules besides the human's understanding on the rules themselves. 
When constructing the RAVEN dataset, four types of rules could be applied row-wisely on the $3 \times 3 $ image matrix: \textit{Constant}, \textit{Progression}, \textit{Arithmetic} and \textit{Distribute\_Three}. Those rules can be defined as follow:

\noindent \textbf{Definition 1} \textit{A question matrix $\boldsymbol{A}^q$  satisfies Constant Rule iff $\boldsymbol{a}^q_3=\boldsymbol{a}^q_1=\boldsymbol{a}^q_2$. }

\noindent \textbf{Definition 2} \textit{A question matrix $\boldsymbol{A}^q$  satisfies Progression Rule iff $\boldsymbol{a}^q_3=2\boldsymbol{a}^q_2-\boldsymbol{a}^q_1$. Note that $\boldsymbol{a}^q_1\ne \boldsymbol{a}^q_2$, otherwise it becomes a Constant Rule. }

\noindent \textbf{Definition 3} \textit{A question matrix $\boldsymbol{A}^q$  satisfies Arithmetic Rule iff $\boldsymbol{a}^q_3=\boldsymbol{a}^q_1\pm \boldsymbol{a}^q_2$. }

\noindent \textbf{Definition 4} \textit{A question matrix $\boldsymbol{A}^q$  satisfies Distribute\_Three Rule iff $\boldsymbol{a}^q_2= \boldsymbol{S} \boldsymbol{a}^q_1$ (observed) and $\boldsymbol{a}^q_3= \boldsymbol{S} \boldsymbol{a}^q_2$, where $\boldsymbol{S}=\begin{bmatrix}[1]
	0 & 1 & 0 \\
	0 & 0 & 1 \\
	1 & 0 & 0 \\
	\end{bmatrix} \text{ or } \begin{bmatrix}[1]
	0 & 0 & 1 \\
	1 & 0 & 0 \\
	0 & 1 & 0 \\
	\end{bmatrix} $.}

Provided the definitions of rules above, in the rest of the section, we aim to prove that a set of reasoning rules can be obtained using Eqn. (\ref{eq:ls}), which correspond to these four types of rules, as outlined in \textbf{Theorem 1}.

\noindent \textbf{Theorem 1} \textit{By solving the RAVEN problems using Eqn. (\ref{eq:ls}), four general groups of parameterized reasoning functions can be yielded as Eqn.~(\ref{eq:cst_2}), (\ref{eq:pgs_2}), (\ref{eq:arm_2}) and (\ref{eq:dis_2}), which correspond to rules: Constant, Progression, Arithmetic and Distribute\_Three, respectively.}

To prove \textbf{Theorem 1}, we need to prove the sufficient and necessary conditions of four independent groups of reasoning rules as stated in \textbf{Lemma 1-4}, which relate to \textit{Constant}, \textit{Progression}, \textit{Arithmetic} and \textit{Distribute\_Three}, respectively. 

\noindent \textbf{Lemma 1} \textit{The Constant Rule can be equivalently represented as a set of parameterized rules as Eqn. (\ref{eq:cst_2}) obtained using Eqn. (\ref{eq:ls}). }

\noindent \textbf{Proof.} 
If $\boldsymbol{A}^q$ satisfies the \textit{Constant} rule, based on \textbf{Definition 1}, 
\begin{align} \label{eq:lemma1}
	\boldsymbol{A}^q_{1,2}=\begin{bmatrix}
		\boldsymbol{a}^q_{1} & \boldsymbol{a}^q_1 \\
	\end{bmatrix}, \boldsymbol{a}^q_3=\boldsymbol{a}^q_1. 
\end{align}

Substitute Eqn. (\ref{eq:lemma1}) into Eqn. (\ref{eq:ls}), with mentioned matrix pseudo-inverse $({\boldsymbol{A}^q_{1,2}}^{\top} \boldsymbol{A}^q_{1,2})^{\dagger}$, we can get: 
\begin{align}
	\hat{\boldsymbol{\theta}}_{\text{LS}}^{\textit{C}} &= ({\boldsymbol{A}^q_{1,2}}^\top \boldsymbol{A}^q_{1,2})^{-1} {\boldsymbol{A}^q_{1,2}}^\top \boldsymbol{a}^q_3 = \begin{bmatrix}
		0.5 \\ 0.5
	\end{bmatrix}, \nonumber \\
	\hat{\boldsymbol{\phi}}^{\textit{C}} &= \boldsymbol{a}^q_3 - \boldsymbol{A}^q_{1,2} \hat{\boldsymbol{\theta}}_{\text{LS}}^{\textit{C}} = \boldsymbol{0}. \nonumber
\end{align}

The parameterized function for \textit{Constant} Rule is hence: 
\begin{align}\label{eq:cst_2}
	\boldsymbol{a}^q_3 = \boldsymbol{A}^q_{1,2} \cdot \begin{bmatrix}
		0.5 \\ 0.5
	\end{bmatrix}. 
\end{align}

It is trivial to prove that given Eqn. (\ref{eq:cst_2}) and observed $\boldsymbol{a}^q_1$ and $\boldsymbol{a}^q_2$, $\boldsymbol{A}^q$ satisfies the \textit{Constant} Rule. $\square$

\noindent \textbf{Lemma 2} \textit{The Progression Rule can be equivalently represented as a set of parameterized rules as Eqn. (\ref{eq:pgs_2}) obtained using Eqn. (\ref{eq:ls}). }

\noindent \textbf{Proof.} 
If $\boldsymbol{A}^q$ satisfies the \textit{Progression} rule, based on \textbf{Definition 2}, 
\begin{align} \label{eq:lemma3}
	\boldsymbol{A}^q_{1,2}=\begin{bmatrix}
		\boldsymbol{a}^q_{1} & \boldsymbol{a}^q_2 \\
	\end{bmatrix}, \boldsymbol{a}^q_3=2\boldsymbol{a}^q_2-\boldsymbol{a}^q_1.
\end{align}

Substitute Eqn. (\ref{eq:lemma3}) into Eqn. (\ref{eq:ls}), we can get: 
\begin{align}
	\hat{\boldsymbol{\theta}}_{\text{LS}}^{\textit{P}} &= ({\boldsymbol{A}^q_{1,2}}^\top \boldsymbol{A}^q_{1,2})^{-1} {\boldsymbol{A}^q_{1,2}}^\top \boldsymbol{a}^q_3 = \begin{bmatrix}
		-1 \\ 2
	\end{bmatrix}, \nonumber\\
	\hat{\boldsymbol{\phi}}^{\textit{P}} &= \boldsymbol{a}^q_3 - \boldsymbol{A}^q_{1,2} \hat{\boldsymbol{\theta}}_{\text{LS}}^{\textit{P}} = \boldsymbol{0}. \nonumber 
\end{align}

The parameterized function for \textit{Progression} Rule is hence: 
\begin{align}\label{eq:pgs_2}
	\boldsymbol{a}^q_3 = \boldsymbol{A}^q_{1,2} \cdot \begin{bmatrix}
		-1 \\ 2
	\end{bmatrix}.
\end{align}
It is trivial to prove that given Eqn. (\ref{eq:pgs_2}), $\boldsymbol{A}^q$ satisfies the \textit{Progression} Rule. $\square$

\noindent \textbf{Lemma 3} \textit{The Arithmetic Rule can be equivalently  represented as a set of parameterized rules as Eqn. (\ref{eq:arm_2}) obtained using Eqn. (\ref{eq:ls}). }

\noindent \textbf{Proof.} 
If $\boldsymbol{A}^q$ satisfies the Arithmetic rule, based on \textbf{Definition 3}, 
\begin{align} \label{eq:lemma5}
	\boldsymbol{A}^q_{1,2}=\begin{bmatrix}
		{\boldsymbol{a}}^q_{1} & \boldsymbol{a}^q_2 \\
	\end{bmatrix}, \boldsymbol{a}^q_3=\boldsymbol{a}^q_1 \pm \boldsymbol{a}^q_2. 
\end{align}

Substitute Eqn. (\ref{eq:lemma5}) into Eqn. (\ref{eq:ls}), we can get: 
\begin{align}
	\hat{\boldsymbol{\theta}}_{\text{LS}}^{\textit{A}} &= ({\boldsymbol{A}^q_{1,2}}^\top \boldsymbol{A}^q_{1,2})^{-1} {\boldsymbol{A}^q_{1,2}}^\top \boldsymbol{a}^q_3 = \begin{bmatrix}
		1 \\ \pm 1
	\end{bmatrix}, \nonumber\\
	\hat{\boldsymbol{\phi}}^{\textit{A}} &= \boldsymbol{a}^q_3 - \boldsymbol{A}^q_{1,2} \hat{\boldsymbol{\theta}}_{\text{LS}}^{\textit{A}} = \boldsymbol{0}. \nonumber 
\end{align}

The parameterized function for \textit{Arithmetic} Rule is hence:
\begin{align}\label{eq:arm_2}
	\boldsymbol{a}^q_3 = \boldsymbol{A}^q_{1,2} \cdot \begin{bmatrix}
		1 \\ \pm 1
	\end{bmatrix}.
\end{align}
It is trivial to prove that given Eqn. (\ref{eq:arm_2}), $\boldsymbol{A}^q$ satisfies the \textit{Arithmetic} Rule. $\square$

\noindent \textbf{Lemma 4} \textit{The Distribute\_Three Rule can be equivalently represented as a set of parameterized rules as Eqn. (\ref{eq:dis_2}) obtained using Eqn. (\ref{eq:ls}). }

\noindent \textbf{Proof.} 
If $\boldsymbol{A}^q$ satisfies the Distribute\_Three rule, based on \textbf{Definition 4}, 
\begin{align} \label{eq:lemma7}
	& \boldsymbol{A}^q_{1,2}=\begin{bmatrix}
		{\boldsymbol{a}}^q_{1} & \boldsymbol{S} \boldsymbol{a}^q_1 \\
	\end{bmatrix}, \boldsymbol{a}^q_3 = \boldsymbol{S} \boldsymbol{a}^q_2 = \boldsymbol{S} (\boldsymbol{S} \boldsymbol{a}^q_1), \nonumber \\
	& \boldsymbol{S}=\begin{bmatrix}
		0 & 1 & 0 \\
		0 & 0 & 1 \\
		1 & 0 & 0 \\
	\end{bmatrix} \text{ or } \begin{bmatrix}
		0 & 0 & 1 \\
		1 & 0 & 0 \\
		0 & 1 & 0 \\
	\end{bmatrix}.
\end{align}

Substitute Eqn. (\ref{eq:lemma7}) into Eqn. (\ref{eq:ls}), we can get: 
\begin{align}
	\hat{\boldsymbol{\theta}}_{\text{LS}}^{\textit{D}} &= ({\boldsymbol{A}^q_{1,2}}^\top \boldsymbol{A}^q_{1,2})^{-1} {\boldsymbol{A}^q_{1,2}}^\top \boldsymbol{a}^q_3 = \begin{bmatrix}
		\frac{s}{p+s} \\
		\frac{s}{p+s}
	\end{bmatrix}, \nonumber \\
	\hat{\boldsymbol{\phi}}^{\textit{D}} &= \boldsymbol{a}^q_3 - \boldsymbol{A}^q_{1,2} \hat{\boldsymbol{\theta}}_{\text{LS}}^{\textit{D}} \nonumber \\
	&= \left(\boldsymbol{S}^{\top}- \frac{s}{p+s}(\boldsymbol{I}_3+\boldsymbol{S})\right)\boldsymbol{a}^q_1, \nonumber
\end{align}
where $p={\boldsymbol{a}^q_1}^\top \boldsymbol{a}^q_1$, $s= {\boldsymbol{a}^q_1}^\top \boldsymbol{S} \boldsymbol{a}^q_1$. 
The parameterized function for the \textit{Distribute\_Three} Rule is hence: 
\begin{align}\label{eq:dis_2}
	\boldsymbol{a}^q_3 = \boldsymbol{A}^q_{1,2} \cdot \begin{bmatrix}
		\frac{s}{p+s} \\
		\frac{s}{p+s}
	\end{bmatrix} + \left(\boldsymbol{S}^{\top}-
	\frac{s}{p+s}(\boldsymbol{I}_3+\boldsymbol{S})\right)\boldsymbol{a}^q_1.
\end{align}
It is trivial to prove that given Eqn. (\ref{eq:dis_2}), $\boldsymbol{A}^q$ satisfies \textit{Distribute\_Three} Rule. $\square$

\section{RPM-like Video Prediction Dataset}
\label{sec4}
A RPM-like Video Prediction (RVP) dataset is constructed to evaluate the proposed method in a practical real-world application. Video prediction relies on understanding the spatio-temporal information of historical frames. Traditionally, a set of previous video frames are needed to predict the next frame \cite{oprea2020review}, while in this paper, the future frame at a time interval is predicted based on two previous distant frames only. 

\subsection{Dataset Construction}
The RVP dataset is constructed based on the UA-DETRAC dataset for object detection and tracking \cite{wen2020ua}, consisting of 100 challenging videos captured from real-world traffic scenes. The videos are selected from over 10 hours of monitoring videos acquired by a Canon EOS 550D camera at 24 different locations, which represent various traffic patterns and conditions including urban highway, traffic crossings and T-junctions. The UA-DETRAC dataset consists of over 140,000 frames with rich annotations such as illumination, vehicle type, occlusion, truncation ratio, and vehicle bounding boxes.

\begin{figure*}[ht]
	\centering
	\subfigure[A panel of question frames extracted from real-world videos.]{
		\begin{minipage}[t]{0.8\linewidth}
			\centering
			\includegraphics[width=1\linewidth]{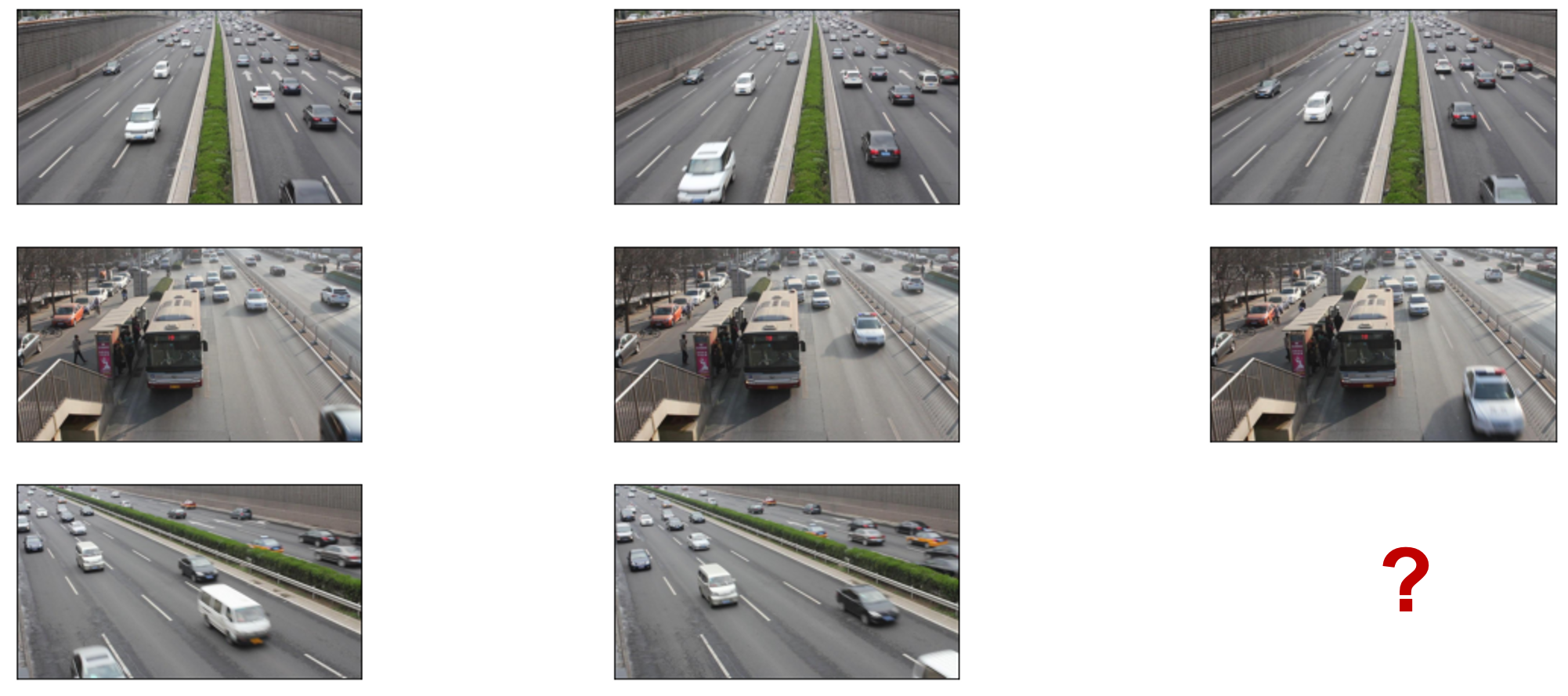}
		\end{minipage}
	}\\
	\subfigure[Candidate answer frames extracted from real-world videos.]{
		\begin{minipage}[t]{0.8\linewidth}
			\centering
			\includegraphics[width=1\linewidth]{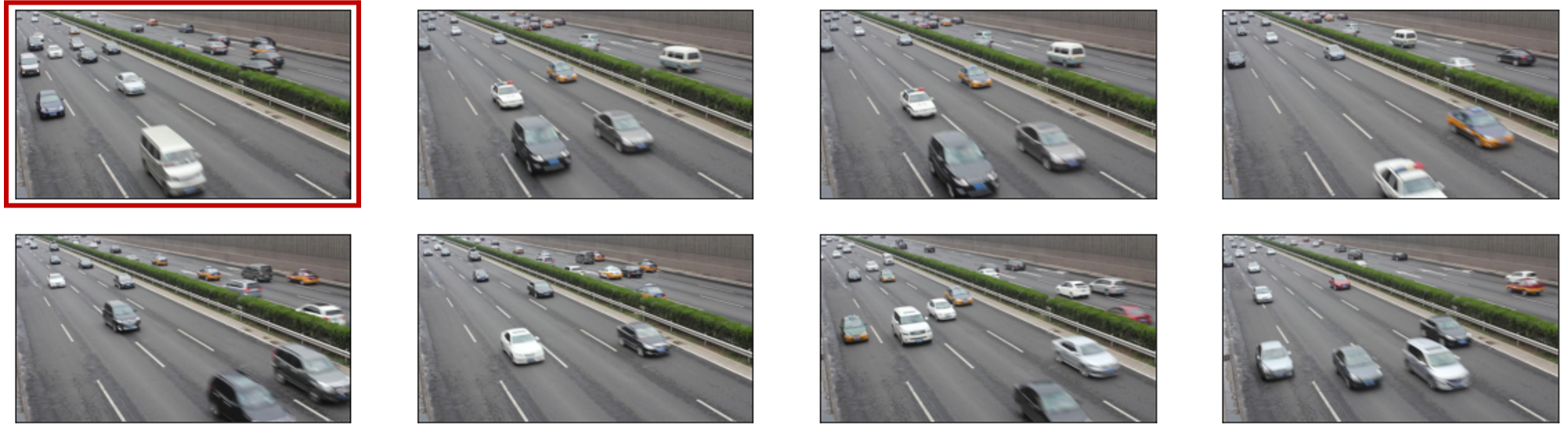}
		\end{minipage}
	}
	\centering
	\caption{An example of the newly developed RPM-like Video Prediction dataset used for street scene understanding and video prediction. In (a), the three rows represent three different video sequences extracted from the UA-DETRAC dataset \cite{wen2020ua}, and the frames are extracted at a predefined time interval. Based on the reasoning rules learned from the first two rows, the target is to select the proper future frame in the third row based on the spatio-temporal information extracted from the first two frames. The correct one is marked in red.}
	\label{fig:RVP}
\end{figure*}

The RVP dataset is constructed under the RPM framework, i.e., the future frame is predicted based on the two historical frames sampled at an interval of 15 frames, using the underlying reasoning rules learned from the first two groups of frames sampled similarly from other videos. Such a problem formulation is conceptually more challenging than the classical video prediction tasks, as only two historical frames sampled every 15 frames are given. To simplify the question, the target is to select the correct one from the eight possible answers, where in addition to the correct frame, the other seven candidate answers are randomly sampled at least 15 frames away from the correct frame. One sample question is shown in Fig.~\ref{fig:RVP}. The developed RVP dataset consists of 3,000 questions for video prediction. Each sample consists of 8 question frames and 8 candidate frames. In total, the dataset consists of 48,000 frames. The developed RVP dataset contains visually complex scenes including environmental backgrounds, multiple vehicle objects and street sights.

\subsection{Adaptation of Proposed TRIVR to RVP Dataset}

\begin{figure}[ht]
	\centering
	\subfigure[Original frame]{
		\begin{minipage}[t]{0.457\linewidth}
			\centering
			\includegraphics[width=1\linewidth]{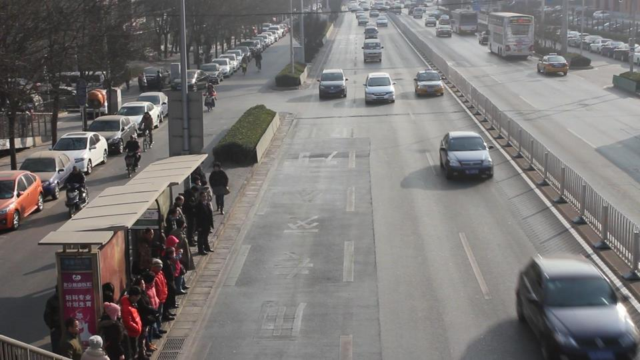}
		\end{minipage}
	}
	\subfigure[Vehicle detection]{
		\begin{minipage}[t]{0.457\linewidth}
			\centering
			\includegraphics[width=1\linewidth]{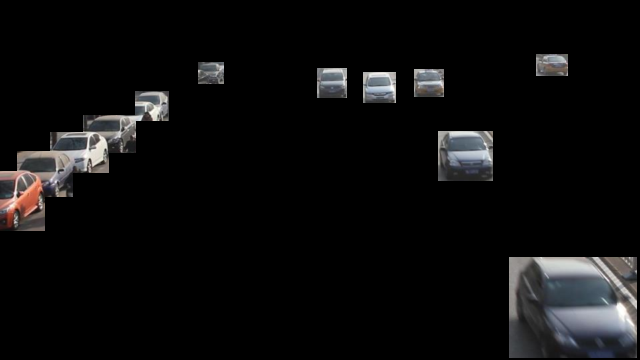}
		\end{minipage}
	}
	\subfigure[Vehicle matching using SIFT features.]{
		\begin{minipage}[t]{0.95\linewidth}
			\centering
			\includegraphics[width=1\linewidth]{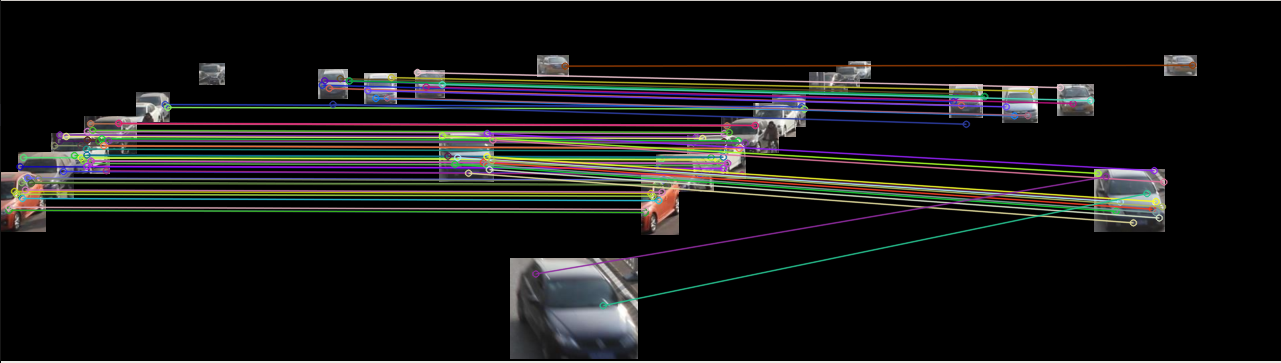}
		\end{minipage}
	}
	\centering
	\caption{Overview of the perception module of proposed TRIVR for the RVP dataset. Objects in input frames are firstly detected using Mask R-CNN. After masking off the background, the corresponding pair of objects are matched using SIFT features. Then the derived attributes such as \textit{Position}, \textit{Size} and \textit{Color} of objects can be used in the subsequent logic reasoning.}
	\label{fig:perception}
\end{figure}

As the image scene of the RVP dataset is much more complicated than that of the common RPM datasets, the perception module of the proposed TRIVR is adapted to the RVP dataset. First of all, all the moving objects in each of the frames are detected using Mask R-CNN \cite{he2017mask}, a state-of-the-art object detector. The Mask R-CNN is robust to recognize almost all moving objects in frames, as shown in Fig.~\ref{fig:perception}(b). However, identifying the same object across frames remains a challenge. To tackle this, the scale invariant feature transform (SIFT) \cite{lowe2004distinctive} is utilized to match objects across frames, as shown in Fig.~\ref{fig:perception}(c). After the pair of objects are matched across different frames, the reasoning module is then applied on the derived attributes such as \textit{Position}, \textit{Size} and \textit{Color} of detected objects across frames, to derive the spatio-temporal reasoning rules. Thanks to the proposed two-step formulation, the adaptation on the perception module does not significantly impact the subsequent reasoning module.

\section{Experimental Results}
\label{sec5}
RAVEN \cite{zhang2019raven} and I-RAVEN \cite{hu2021stratified} datasets are used to evaluate the proposed TRIVR and other methods. Experiments are also conducted on the shrunk version of the two datasets. Furthermore, the proposed method and state-of-the-art models are evaluated on the developed RVP dataset for video prediction.

\subsection{Experimental Settings}
\subsubsection{Datasets}
Two benchmark datasets, RAVEN \cite{zhang2019raven} and I-RAVEN \cite{hu2021stratified} datasets, are used in the experiments. In addition, to show that the proposed method requires much fewer training samples, a portion of training data of RAVEN and I-RAVEN datasets are used whereas the test data and validation data remain the same. These datasets are referred as Shrunk-RAVEN dataset.

\noindent \textbf{RAVEN:} 
The size of question sets in the RAVEN dataset \cite{zhang2019raven} is 70,000. The dataset is randomly split into 10 folds, with 6 folds for training, i.e. 42,000 training samples, and 2 folds for validation and 2 folds for testing respectively. Each question contains 8 question images, arranged as a $3 \times 3$ image matrix with the last one missing, and 8 candidate answers. In the RAVEN dataset, given the first 8 images of the question, the objective is to derive the reasoning rules from the first 8 images and choose one image from the 8 candidate images as the correct one. The candidate answers are generated by permutation from the ground-truth answer image, and each permutated image is derived by randomly shifting one attribute value. The dataset is equally distributed into 7 problem configurations, i.e. \textit{Center}, \textit{2$\times$2 Grid}, \textit{3$\times$3 Grid}, \textit{Left-Right}, \textit{Up-Down}, \textit{Out-InCenter} and \textit{Out-InGrid}. Each question contains 6 visual attributes (\textit{Angle}, \textit{Number}, \textit{Position}, \textit{Type}, \textit{Size} and \textit{Color}) and 4 underlying rules (\textit{Constant}, \textit{Progression}, \textit{Distribute\_Three} and \textit{Arithmetic}). 

Attribute \textit{Angle} is an irrelevant attribute in solving the RPM problem, whereas other attributes may follow one of the underlying reasoning rules. In configuration \textit{Center}, \textit{Left-Right}, \textit{Up-Down} and \textit{Out-InCenter}, attribute \textit{Number} and \textit{Position} are the same for all questions in the same configuration. In configuration \textit{2$\times$2 Grid}, \textit{3$\times$3 Grid} and \textit{Out-InGrid}, either \textit{Number} or \textit{Position} is selected as the underlying construction rule, which implies that the other one becomes an irrelevant attribute, and hence complicates the problem, as misleading rules may be generated from these irrelevant attributes. 

Besides irrelevant attributes, Zhang \etal \cite{zhang2019raven} mentioned that they introduced extra \textit{noise} information to problem panels as \textit{Uniformity}. It refers to randomly changing portions of entities in question samples. See Fig.~\ref{example} for example, the \textit{Color} values vary freely but \textit{Number}, \textit{Type} and \textit{Size} attributes follow the underlying rules. Specifically, the question panel was originally set with \textit{Constant} rule in \textit{Color} attribute, but it was enshadowed with random color values. Therefore, the \textit{Color} attribute may mislead the reasoning process and hence the problem becomes more challenging.

\begin{figure}[h]
	\centering
	\includegraphics[width=.6\linewidth]{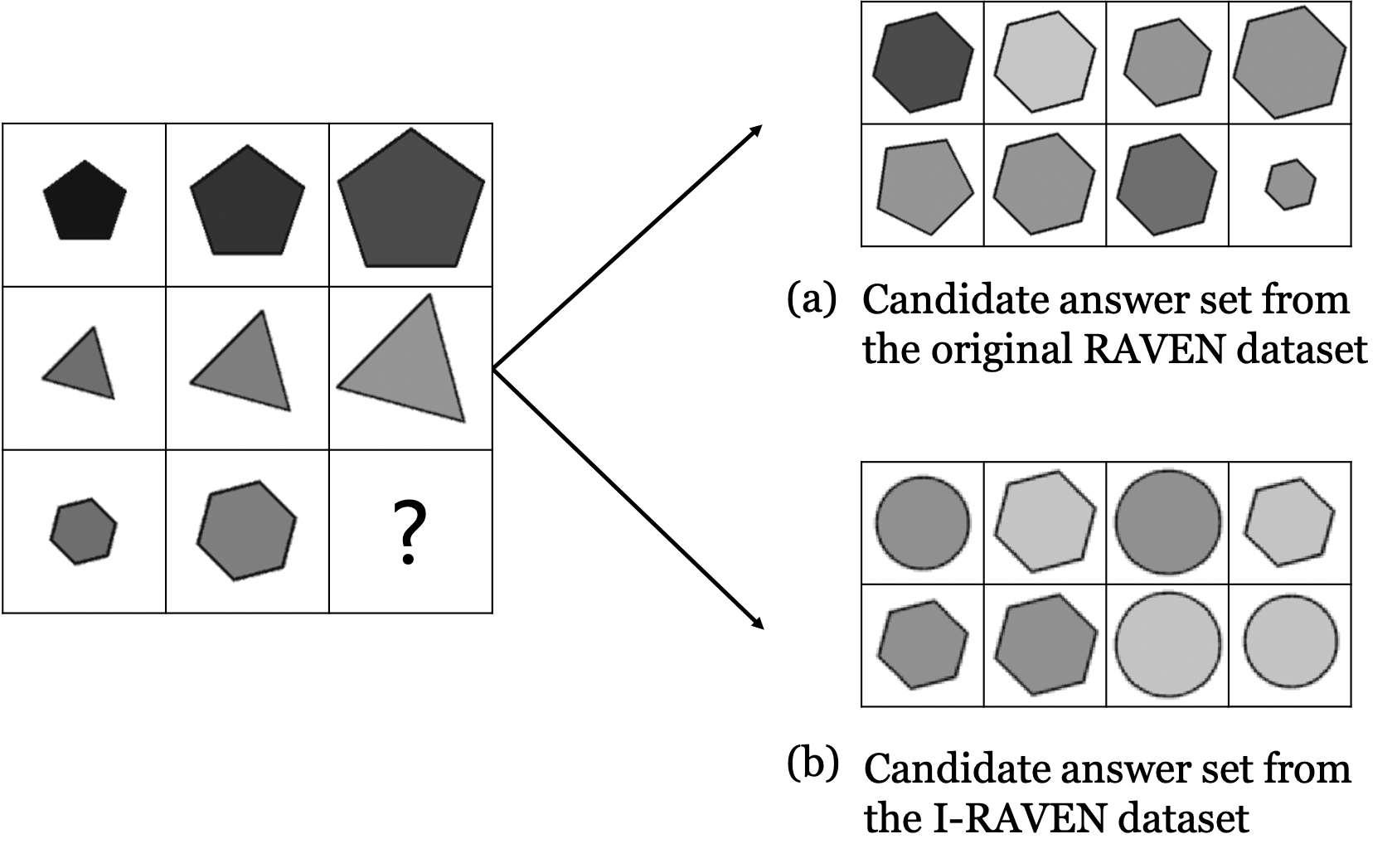}
	\caption{Different candidate answer sets by the RAVEN dataset \cite{zhang2019raven} and the I-RAVEN dataset \cite{hu2021stratified}. One may aggregate the correct answer by selecting the most common attribute from the candidate answers in the original RAVEN dataset, while not feasible in the I-RAVEN dataset.}
	\label{R_vs_B}
\end{figure}

\noindent \textbf{Impartial-RAVEN (I-RAVEN):} 
The original RAVEN dataset may lead to a shortcut that the aggregation of most common values for each attribute could be the correct answer \cite{hu2021stratified}. To address this problem, Hu \etal \cite{hu2021stratified} designed an I-RAVEN dataset, by changing the permutation scheme of the candidate answers. Instead of random permutation of one attribute at a time in the RAVEN dataset, the negative candidate answers of the I-RAVEN dataset are generated by hierarchically permuting one attribute of the ground-truth answer in three iterations. In each iteration, two child nodes are generated, where one node remains the same with the parent node and the other permutes one attribute. An example is shown in Fig. \ref{R_vs_B}(b). Other than this, the same evaluation protocol as in the original RAVEN dataset is used. 

\noindent \textbf{Shrunk-RAVEN: }
One of the major advantages of the proposed approach is that one reasoning rule can be derived from each sample question, thanks to the proposed two-step framework and ``2+1'' formulation, while existing approaches need many more samples. The proposed TRIVR only requires a minimal amount of training samples covering the possible reasoning rules. To verify this, we conduct a series of experiments on the shrunk-RAVEN datasets, where only a portion of the original dataset is randomly selected for training, while the testing samples of the original dataset remain unchanged. As shown later, the proposed approach achieves a competitive performance by utilizing only approximately 1/64 of the original training data. Similar experiments are carried out on both the RAVEN and I-RAVEN datasets.

\subsubsection{Compared Methods}
The default image size of $80 \times 80$ is used to avoid unexpected effect of image size to the reasoning accuracy. The proposed method is compared with CNN-based models, which mostly utilize the ``8+1'' formulation, including LSTM, WReN \cite{santoro2018measuring}, CNN, ResNet-18 \cite{he2016deep}, ResNet-18+DRT \cite{zhang2019raven}, LEN \cite{zheng2019abstract}, CoPINet \cite{zhang2019learning} and SRAN \cite{hu2021stratified}. These approaches are recently published in reputable journals or conferences. The experimental results are mainly referred from \cite{zhang2019learning} and \cite{hu2021stratified}. 

\noindent \textbf{LSTM} Zhang \etal \cite{zhang2019raven} extracted image features using a CNN, followed by an LSTM network, and added a two-layer MLP at the end for prediction.

\noindent \textbf{WReN} Wild Relation Network \etal \cite{santoro2018measuring} utilizes the Relation Network \cite{santoro2017simple} to calculate the probability of each candidate answer to be the correct one. 

\noindent \textbf{CNN} This method contains a four-layer CNN for feature extraction and a two-layer MLP with a softmax layer for prediction, as outlined in \cite{zhang2019raven}. 

\noindent \textbf{ResNet-18} The original ResNet-18 is utilized as the feature extractor and a two-layer MLP with a softmax layer is used for prediction, as outlined in \cite{zhang2019raven}. 

\noindent \textbf{ResNet-18+DRT} Zhang \etal \cite{zhang2019raven} enhanced the original ResNet-18 with a Dynamic Residual Tree (DRT) module, which takes the features from the lower layers into the leaf nodes of DRT and gradually updates using the ReLU function following the tree structure, with a residual structure to add the original input to form the output.  

\noindent \textbf{LEN} Logic Embedding Network (LEN) developed by Zheng \etal \cite{zheng2019abstract} assembles the possible candidate answer embeddings to the 8 question panel embeddings, and calculates scores for all possible combinations ($C_9^3=84$). The model outputs the predicted answer with the highest score. 

\noindent \textbf{CoPINet} Zhang \etal \cite{zhang2019learning} designed Contrastive Perceptual Inference Network (CoPINet), which achieves the previously published best results on the RAVEN dataset. It models the probability of each candidate answer by applying a contrasting module on a perception module, with the question panel and each one of the candidate answers as the input. 

\noindent \textbf{SRAN} Hu \etal \cite{hu2021stratified} developed Stratified Rule-Aware Network (SRAN), which achieves the previously published best results on the I-RAVEN dataset. It utilizes a hierarchical rule embedding module and a gated embedding fusion module to output the rule embedding given two row sequence. 

\subsection{Comparisons to State-of-the-art Models on RPMs}

\subsubsection{Performance on RAVEN}
The proposed method is compared to state-of-the-art models \cite{santoro2017simple,zhang2019raven,zheng2019abstract,zhang2019learning}, and the results are summarized in Table \ref{tab:raven}. It can be seen that in \textit{Center}, \textit{Left-Right}, \textit{Up-Down} and \textit{Out-InCenter} configurations, the proposed method achieves almost perfect accuracy. Attributes \textit{N} and \textit{P} are the same for these configurations, and hence have the minimal disturbance on the reasoning process. Thanks to the proposed two-step framework and ``2+1'' formulation, one rule can be extracted from each question panel for each attribute to form the rule pool, which helps the proposed TRIVR in the testing phase to solve questions based on the derived rule. 
\begin{table*}[h]
	\setlength{\belowcaptionskip}{5mm}
	\centering
	\caption{The comparisons of reasoning accuracy on the RAVEN dataset. All the results for other models are referred from \cite{zhang2019learning}. The proposed TRIVR outperforms the previously published best approach, CoPINet \cite{zhang2019learning}, by 2.2\% on average. }
	
	\resizebox{\linewidth}{!}{
		\begin{tabular}{l|c|ccccccc}
			\toprule
			\textbf{Methods} & \textit{\textbf{Acc}} & \textit{Center} & \textit{2$\times$2 Grid} & \textit{3$\times$3 Grid} & \textit{Left-Right} & \textit{Up-Down} & \textit{Out-InCenter} & \textit{Out-InGrid}  \\
			\midrule
			LSTM & 13.1\% & 13.2\% & 14.1\% & 13.7\% & 12.8\% & 12.4\% & 12.2\% & 13.0\% \\
			WReN \cite{santoro2018measuring} & 34.0\% & 58.4\% & 38.9\% & 37.7\% & 21.6\% & 19.7\% & 38.8\% & 22.6\% \\
			CNN & 37.0\% & 33.6\% & 30.3\% & 33.5\% & 39.4\% & 41.3\% & 43.2\% & 37.5\% \\
			ResNet & 53.4\% & 52.8\% & 41.9\% & 44.3\% & 58.8\% & 60.2\% & 63.2\% & 53.1\% \\
			ResNet+DRT \cite{zhang2019raven} & 59.6\% & 58.1\% & 46.5\% & 50.4\% & 65.8\% & 67.1\% & 69.1\% & 60.1\% \\
			LEN \cite{zheng2019abstract} & 72.9\% & 80.2\% & 57.5\% & 62.1\% & 73.5\% & 81.2\% & 84.4\% & 71.5\% \\
			CoPINet \cite{zhang2019learning} & 91.4\% & 95.1\% & 77.5\% & 78.9\% & 99.1\% & 99.7\% & 98.5\% & \textbf{91.4\%} \\
			\textbf{Proposed TRIVR} & \bf{93.6\%} & \bf{99.9\%} & \textbf{86.0\%} & \textbf{80.9\%} & \bf{99.9\%} & \bf{100.0\%} & \bf{99.8\%} & 88.9\% \\
			\bottomrule
		\end{tabular}
	}
	\label{tab:raven}
\end{table*}

Although for \textit{Center}, \textit{Left-Right} and \textit{Out-InCenter} configurations, the proposed method achieves a nearly perfect accuracy, there are still some failure cases. One remaining challenge is that attributes of the first two rows for some testing question panels ${\boldsymbol{I}^q}^\prime$ are identical so that it is not sufficient to determine any concrete rule, which may lead to failure cases. In \textit{2}$\times$\textit{2} \textit{Grid}, \textit{3}$\times$\textit{3} \textit{Grid} and \textit{Out-InGrid} configurations, the accuracy is relatively low compared to other configurations, which is mainly due to the existence of irrelevant attributes in these configurations. It is difficult to detect irrelevant attributes using one sample question only. CoPINet \cite{zhang2019learning} aggregates the reasoning rules based on a set of samples, which could be the reason why it outperforms the proposed method in the \textit{O-IG} configuration. Other than this configuration, the proposed method outperforms the previously best performed method, CoPINet, on the RAVEN dataset, with an average performance gain of 2.2\%.

\subsubsection{Performance on I-RAVEN}
The proposed method is compared to a variety of models \cite{santoro2017simple,zhang2019raven,zheng2019abstract,zhang2019learning,hu2021stratified}. The results are summarized in Table \ref{tab:i-raven}. As shown in Table \ref{tab:i-raven}, the proposed method consistently and significantly outperforms all the compared approaches. It improves the previously published best results by SRAN \cite{hu2021stratified} from 60.8\% to 95.9\% on average. 
\begin{table*}[ht]
	\centering
	\caption{The comparisons of reasoning accuracy on the I-RAVEN dataset. The results for other models are referred from \cite{hu2021stratified}. The proposed TRIVR improves the performance of previously published best method, SRAN \cite{hu2021stratified}, from 60.8\% to 95.9\% on average. }
	\resizebox{\linewidth}{!}{
	\begin{tabular}{l|c|ccccccc}
		\toprule
		\textbf{Methods} & \textit{\textbf{Acc}} & \textit{Center} & \textit{2$\times$2 Grid} & \textit{3$\times$3 Grid} & \textit{Left-Right} & \textit{Up-Down} & \textit{Out-InCenter} & \textit{Out-InGrid}  \\
		\midrule
		LSTM & 18.9\% & 26.2\% & 16.7\% & 15.1\% & 14.6\% & 16.5\% & 21.9\% & 21.1\% \\
		WReN \cite{santoro2018measuring}  & 23.8\% & 29.4\% & 26.8\% & 23.5\% & 21.9\% & 21.4\% & 22.5\% & 21.5\% \\
		ResNet & 40.3\% & 44.7\% & 29.3\% & 27.9\% & 51.2\% & 47.4\% & 46.2\% & 35.8\% \\
		ResNet+DRT \cite{zhang2019raven}  & 40.4\% & 46.5\% & 28.8\% & 27.3\% & 50.1\% & 49.8\% & 46.0\% & 34.2\% \\
		LEN \cite{zheng2019abstract}  & 41.4\% & 56.4\% & 31.7\% & 29.7\% & 44.2\% & 44.2\% & 52.1\% & 31.7\% \\
		Wild ResNet & 44.3\% & 50.9\% & 33.1\% & 30.8\% & 53.1\% & 52.6\% & 50.9\% & 38.7\% \\
		CoPINet \cite{zhang2019learning}  & 46.1\% & 54.4\% & 36.8\% & 31.9\% & 51.9\% & 52.5\% & 52.2\% & 42.8\% \\
		SRAN \cite{hu2021stratified}  & 60.8\% & 78.2\% & 50.1\% & 42.4\% & 70.1\% & 70.3\% & 68.2\% & 46.3\% \\
		\textbf{Proposed TRIVR} & \bf{95.9\%} & \bf{100.0\%} & \textbf{89.8\%} & \bf{88.3\%} & \bf{99.9\%} & \bf{100.0\%} & \bf{100.0\%} & \bf{93.6\%} \\
		\bottomrule
	\end{tabular}
}
	\label{tab:i-raven}
\end{table*}

In \textit{Center}, \textit{Left-Right}, \textit{Up-Down} and \textit{Out-InCenter} configurations, the proposed method achieves an almost perfect accuracy. Most models suffer from a performance drop transferring between two datasets, even though they might have excellent performance in original RAVEN, e.g. ResNet+DRT (from 59.6\% to 40.4\%), LEN (from 72.9\% to 41.4\%) and CoPINet (from 91.4\% to 46.1\%). One possible reason is that these models might benefit from the mentioned shortcut to the correct answer implicitly, by contrasting attribute information in candidate answers. The proposed method receives consistent improvements compared with the results obtained on the original RAVEN dataset. The I-RAVEN dataset addresses the problem of the shortcut to the candidate answer in the RAVEN dataset and it is regarded as a more balanced and fair benchmark. At the same time, it shrinks the search space as shown in Fig.~\ref{R_vs_B}.
For example, there are four different colors and four different sizes for candidate answers of the RAVEN dataset shown in Fig.~\ref{R_vs_B}(a), 
whereas in the I-RAVEN dataset shown in Fig.~\ref{R_vs_B}(b), all visual attributes are balanced into 2. 
The proposed method works well on both datasets, and it achieves a much better performance despite the performance drop by other methods when moving from the original RAVEN dataset to the I-RAVEN dataset.

\begin{figure}[htbp]
	\centering
	\subfigure[]{
		\begin{minipage}[t]{0.44\textwidth}
			\centering
			\includegraphics[width=0.6\textwidth]{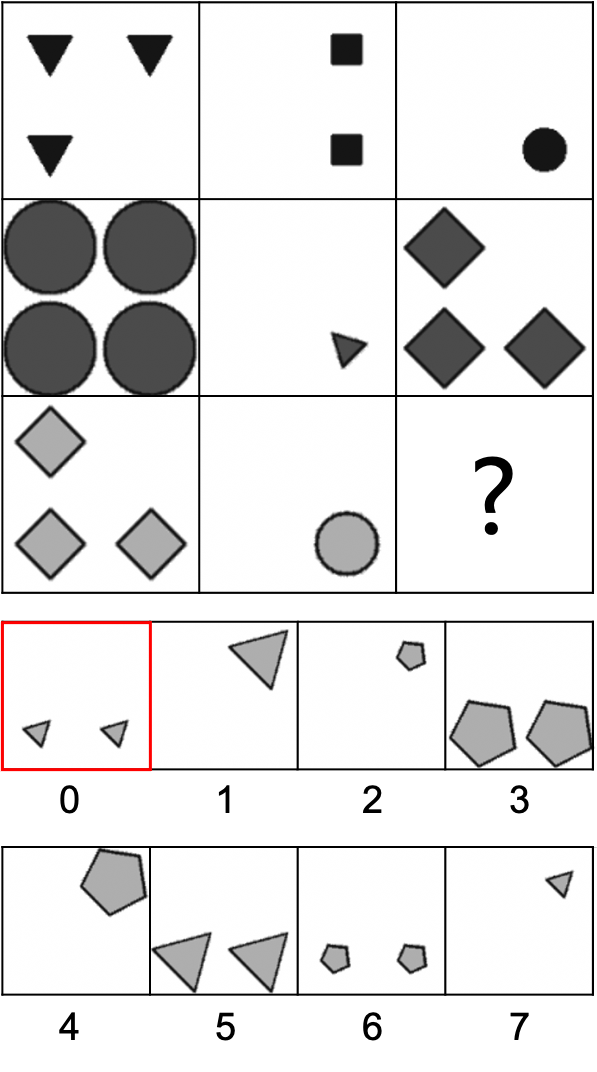}
		\end{minipage}
	}
	\subfigure[]{
		\begin{minipage}[t]{0.44\textwidth}
			\centering
			\includegraphics[width=0.6\textwidth]{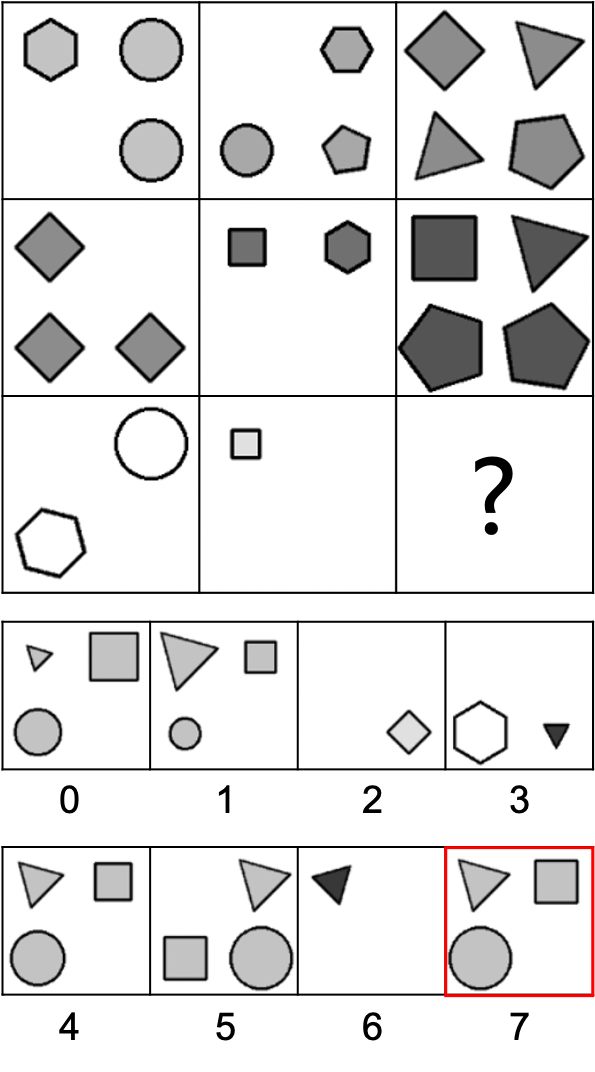}
		\end{minipage}
	}
	
	\caption{Examples of failure cases of SRAN. (a) SRAN predicts Candidate 7, TRIVR predicts Candidate 0. (b) SRAN predicts Candidate 5, TRIVR predicts Candidate 7. Underlying rules: (a) the \textit{Arithmetic} rule on \textit{Number}. (b) the \textit{Arithmetic} rule (addition) on \textit{Position}. }
	\label{fail_cases}
\end{figure}

Existing solution models such as CoPINet and SRAN are built based on the ``8+1'' formulation, which unnecessarily complicates the problem so that they may not be able to discover the underlying rules. As illustrated by the example shown in Fig. \ref{fail_cases}(a), SRAN makes mistakes on analysis of \textit{Number} attribute, while in the proposed method, because of the proposed ``2+1'' formulation and the rule inference process, we could predict the correct answer using the discovered underlying reasoning rule, e.g. the \textit{Arithmetic} rule on \textit{Number}.

\subsubsection{Performance on Shrunk Datasets}
In this experiment, we evaluate the effect of reducing the training dataset size on the reasoning ability of models. The proposed TRIVR is compared with the previously best performed methods, CoPINet \cite{zhang2019learning} on the RAVEN dataset and SRAN \cite{hu2021stratified} on the I-RAVEN dataset, respectively. We use 1/64, 1/32. 1/16, 1/8, 1/4 and 1/2 of the original training dataset, respectively. The results are summarized in Table \ref{shrink-raven} and Table \ref{shrink-i-raven}, respectively. 
\begin{table*}[!ht]
	\setlength{\belowcaptionskip}{5mm}
	\centering
	\caption{The proposed TRIVR is compared with previously best performed method CoPINet \cite{zhang2019learning} using different training dataset sizes on the RAVEN dataset. The proposed method consistently and significantly outperforms CoPINet on all settings.}
	\resizebox{\linewidth}{!}{
		\begin{tabular}{c|c|cccccc}
			\toprule
			\multicolumn{2}{c}{{\textbf{Training }\textbf{set size}}} & 658 & 1,316 & 2,625 & 5,250 & 10,500 & 21,000 \\
			\midrule
			\multirow{8}*{CoPINet \cite{zhang2019learning}} & \textit{Center} & 45.75\% & 47.50\% & 57.70\% & 70.15\% & 98.20\% & 98.50\% \\
			& \textit{2}$\times$\textit{2} \textit{Grid} & 51.00\% & 55.90\% & 65.20\% & 73.85\% & 64.15\% & 78.15\% \\
			& \textit{3}$\times$\textit{3} \textit{Grid} & 53.05\% & 57.65\% & 63.30\% & 68.40\% & 63.05\% & 72.70\% \\
			& \textit{Left-Right} & 56.90\% & 66.25\% & 75.90\% & 88.85\% & 99.10\% & 99.50\% \\
			& \textit{Up-Down} & 55.20\% & 63.75\% & 73.50\% & 85.80\% & 96.45\% & 99.15\% \\
			& \textit{Out-InCenter} & 36.30\% & 53.60\% & 56.55\% & 63.30\% & 85.30\% & 97.50\% \\
			& \textit{Out-InGrid} & 23.15\% & 42.50\% & 42.95\% & 49.50\% & 66.95\% & 78.95\% \\
			& \textit{\textbf{Average}} & 45.91\% & 55.31\% & 62.16\% & 71.41\% & 81.89\% & 89.21\% \\
			\midrule
			\multirow{8}*{TRIVR} & \textit{Center} & 90.50\% & 96.15\% & 98.45\% & 99.60\% & 99.90\% & 99.90\%\\
			& \textit{2}$\times$\textit{2} \textit{Grid} & 77.60\% & 79.80\% & 81.55\% & 82.35\% & 82.65\% & 82.90\% \\
			& \textit{3}$\times$\textit{3} \textit{Grid} & 71.30\% & 72.25\% & 72.75\% & 73.35\% & 73.85\% & 74.30\% \\
			& \textit{Left-Right} & 88.20\% & 97.85\% & 99.40\% & 99.85\% & 99.85\% & 99.85\% \\
			& \textit{Up-Down} & 91.30\% & 95.90\% & 99.15\% & 99.80\% & 100.00\% & 100.00\% \\
			& \textit{Out-InCenter} & 90.45\% & 96.85\% & 98.90\% & 99.75\% & 99.75\% & 99.75\% \\
			& \textit{Out-InGrid} & 81.40\% & 83.20\% & 84.40\% & 84.75\% & 85.00\% & 86.00\% \\
			& \textit{\textbf{Average}} & 84.39\% & 88.86\% & 90.66\% & 91.35\% & 91.57\% & 91.82\% \\
			\bottomrule
		\end{tabular}
	}
	\label{shrink-raven}
\end{table*}

\begin{table*}[h]
	\setlength{\belowcaptionskip}{5mm}
	\centering
	\caption{The proposed TRIVR is compared with previously best performed method SRAN \cite{hu2021stratified} using different training dataset sizes on the I-RAVEN dataset. The proposed method largely outperforms SRAN on all settings.}
	\resizebox{\linewidth}{!}{
		\begin{tabular}{c|c|cccccc}
			\toprule
			\multicolumn{2}{c}{{\textbf{Training }\textbf{set size}}} & 658 & 1,316 & 2,625 & 5,250 & 10,500 & 21,000 \\
			\midrule
			\multirow{8}*{SRAN \cite{hu2021stratified}} & \textit{Center} & 33.20\% & 38.25\% & 51.60\% & 70.05\% & 56.55\% & 78.25\%\\
			& \textit{2}$\times$\textit{2} \textit{Grid} & 22.20\% & 22.55\% & 27.30\% & 26.10\% & 36.65\% & 49.00\%\\
			& \textit{3}$\times$\textit{3} \textit{Grid} & 20.00\% & 19.05\% & 22.05\% & 21.30\% & 30.10\% & 42.75\%\\
			& \textit{Left-Right} & 28.25\% & 30.65\% & 38.55\% & 35.75\% & 48.15\% & 51.05\%\\
			& \textit{Up-Down} & 28.20\% & 29.25\% & 37.80\% & 34.30\% & 47.60\% & 53.05\%\\
			& \textit{Out-InCenter} & 24.80\% & 19.50\% & 29.50\% & 29.80\% & 37.80\% & 60.65\%\\
			& \textit{Out-InGrid} & 19.95\% & 18.40\% & 24.90\% & 24.50\% & 28.85\% & 39.50\%\\
			& \textit{\textbf{Average}} & 25.22\% & 25.40\% & 33.08\% & 34.58\% & 40.81\% & 53.48\% \\
			\midrule
			\multirow{8}*{TRIVR} & \textit{Center} & 98.50\% & 99.60\% & 99.85\% & 99.95\% & 99.95\% & 99.95\%\\
			& \textit{2}$\times$\textit{2} \textit{Grid} & 85.55\% & 87.65\% & 88.85\% & 89.05\% & 89.20\% & 89.60\% \\
			& \textit{3}$\times$\textit{3} \textit{Grid} & 79.05\% & 80.60\% & 81.55\% & 81.95\% & 82.30\% & 83.30\% \\
			& \textit{Left-Right} & 89.65\% & 99.35\% & 99.35\% & 99.75\% & 99.95\% & 99.95\% \\
			& \textit{Up-Down} & 85.70\% & 92.20\% & 99.85\% & 99.85\% & 100.00\% & 100.00\% \\
			& \textit{Out-InCenter} & 87.70\% & 98.65\% & 99.35\% & 99.90\% & 99.90\% & 99.90\%\\
			& \textit{Out-InGrid} & 89.55\% & 90.75\% & 91.50\% & 91.80\% & 91.90\% & 92.00\% \\
			& \textit{\textbf{Average}} & 87.96\% & 92.69\% & 94.34\% & 94.61\% & 94.74\% & 94.96\% \\
			\bottomrule
		\end{tabular}
	}
	\label{shrink-i-raven}
\end{table*}

The proposed method largely outperforms two compared methods on all settings. The smaller the training dataset size, the larger the performance gain. Our approach is robust even using only 1/64 of the training data. While other methods fail to work, the proposed method still retains a high accuracy. The reasoning accuracies of CoPINet and SRAN drop rapidly when the dataset size decreases, indicating that they rely on much more training samples than the proposed method. Thanks to the proposed ``2+1'' formulation that greatly simplifies the problem, the proposed method could extract one reasoning rule per sample question, and hence requires much few samples to extract a pool of reasoning rules. In addition, the proposed TRIVR surpasses human-level (84.41\% on average \cite{zhang2019raven}) using 658 training samples only. 

\begin{figure}[pt]
	\centering
	\subfigure[Results on shrunk \textbf{RAVEN} dataset. ]{
		\begin{minipage}[t]{0.47\textwidth}
			\centering
			\includegraphics[width=0.9\textwidth]{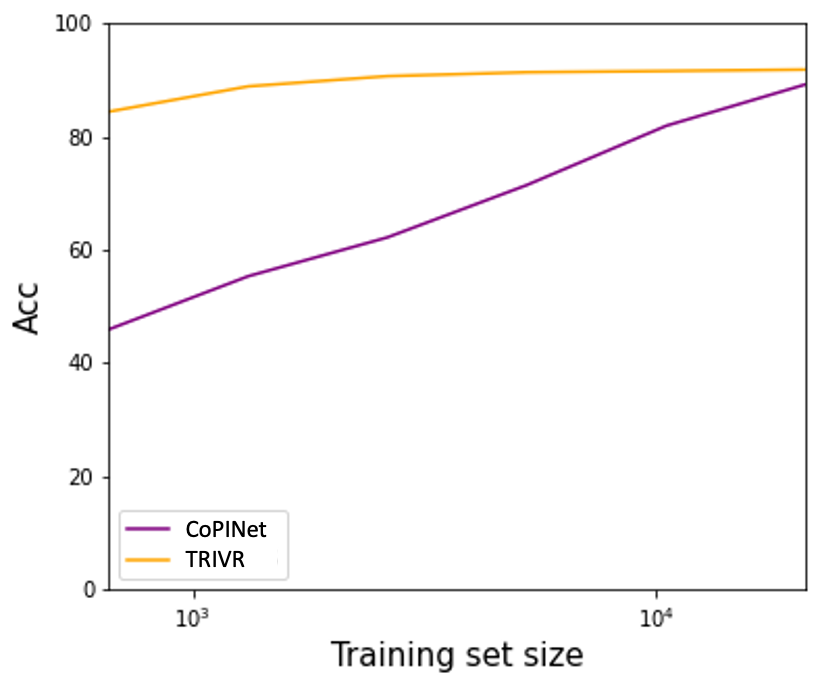}
		\end{minipage}
	}
	\subfigure[Results on shrunk \textbf{I-RAVEN} dataset. ]{
		\begin{minipage}[t]{0.47\textwidth}
			\centering
			\includegraphics[width=0.9\textwidth]{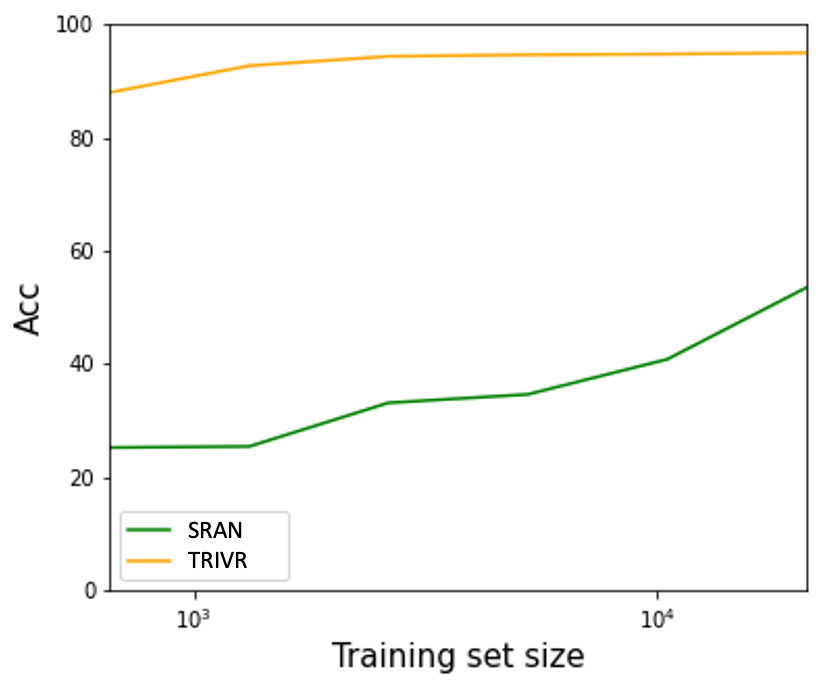}
		\end{minipage}
	}
	\caption{Analysis of influence of training data size on reasoning accuracy for CoPINet and TRIVR on the RAVEN dataset, and SRAN and TRIVR on the I-RAVEN dataset.}
	\label{shrink-plot}
\end{figure}

For a better view, we also plot the mean accuracy across different configurations for the proposed method and the compared methods on the RAVEN dataset and the I-RAVEN dataset for different sizes of the training dataset, as shown in Fig. \ref{shrink-plot}. The experimental results demonstrate the superior reasoning capability of the proposed method over other methods.

\subsection{Comparisons to State-of-the-art Methods on RVP Dataset}
The proposed TRIVR is compared with state-of-the-art methods, CoPINet \cite{zhang2019learning} and SRAN \cite{hu2021stratified}, on the developed RVP dataset to analyze the reasoning ability of the models on real-world video prediction. The RVP dataset has more complicated image scenes than the RAVEN dataset and its variants. We randomly partition the dataset and utilize half of the dataset for training, and the other half for testing. The comparison results are summarized in Table~\ref{tab:result_pred}. 
\begin{table}[htbp]
	\centering
	\normalsize
	\caption{The comparisons of reasoning accuracy on the RVP dataset for video prediction between the proposed method and state-of-the-art methods. The proposed TRIVR outperforms CoPINet \cite{zhang2019learning} by 9.24\% and SRAN \cite{hu2021stratified} by 7.69\%.}
	\begin{tabular}{c|c}
		\hline
		\textit{Models} & \textit{Reasoning Accuracy }\\
		\hline
		CoPINet \cite{zhang2019learning} & 62.34\% \\
		SRAN \cite{hu2021stratified} & 63.87\% \\
		\textbf{Proposed TRIVR} & \textbf{71.56\%} \\
		\hline
	\end{tabular}
	\label{tab:result_pred}
\end{table}

As shown in Table \ref{tab:result_pred}, the proposed method significantly outperforms the state-of-the-art models, CoPINet \cite{zhang2019learning} and SRAN \cite{hu2021stratified}, on the average reasoning accuracy. The performance gain is 9.24\% over CoPINet \cite{zhang2019learning} and 7.69\% over SRAN \cite{hu2021stratified}. 
Thanks to the two-step formulation and ``2+1" framework, the proposed method could adapt to different visual reasoning scenarios. Also benefited from the proposed efficient logic reasoning module, the proposed method significantly outperforms the state-of-the-art methods.

\begin{figure}[h]
	\centering
	\includegraphics[width=0.9\textwidth]{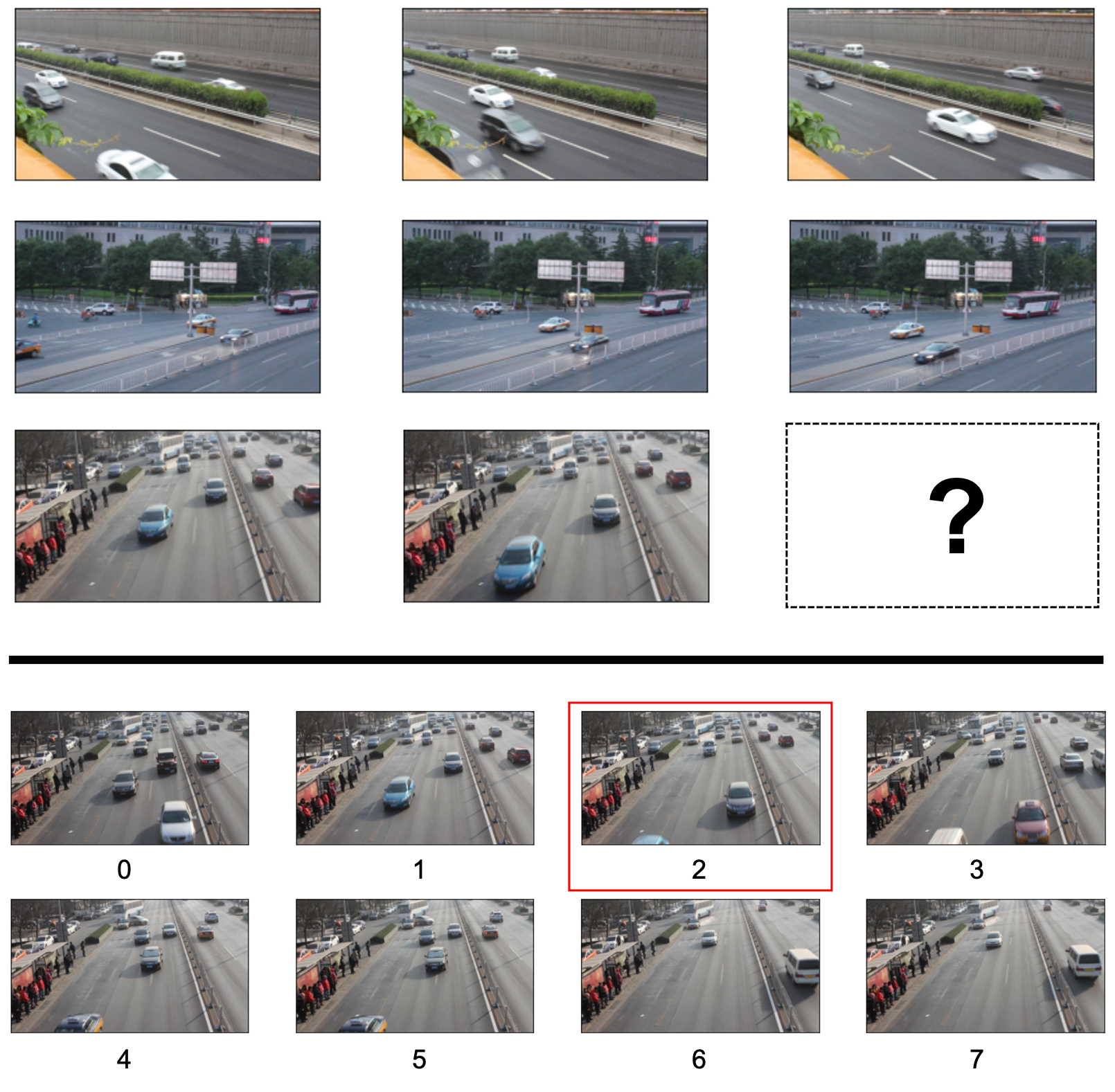}
	\caption{An visual example that the proposed TRIVR can correctly reason over the complex image scenes, while CoPINet \cite{zhang2019learning} can't. The proposed TRIVR correctly predicts Candidate 2 as the correct answer while CoPINet wrongly predicts Candidate 1 as the answer.}
	\label{fail}
\end{figure}



For further illustration, we show one example in Fig. \ref{fail} that the proposed method works well while CoPINet fails. It is indeed difficult to analyze the spatio-temporal relations of objects given only two previous frames at a predefined interval of time, especially when the image scene is complicated, e.g., there are a large number of diversified objects moving in the video sequence. 

\section{Conclusion}
\label{sec6}
In this paper, we introduce an effective solution model for RPM problems. Different from the existing black-box CNN-based solution models, we develop a solution that models the human's solving process by using a two-stage framework. It includes a perception module and a reasoning module to make better use of the developments in both compute vision and logic reasoning societies. 
With the proposed ``2+1'' formulation, RPM problems are solved in an elegant and efficient way by solving small-scale regression problems. The proposed solution could yield human-understandable reasoning rules that previous methods could hardly achieve. An RVP dataset is constructed under the RPM framework to evaluate the visual reasoning methods on scene understanding of practical real-world traffic data. A series of experiments are performed on the RAVEN dataset, RAVEN variants and the RVP dataset to evaluate the robustness and generalization of the proposed model. The proposed method outperforms the state-of-the-art models significantly and consistently on all datasets.




\vspace{-2mm}
\section*{Acknowledgment}
This work was supported in part by the National Natural Science Foundation of China under Grant 72071116, and in part by the Ningbo Municipal Bureau Science and Technology under Grant 2019B10026.

\vspace{-2mm}
\section*{Reference}
{\small
\bibliographystyle{elsarticle-num-names} 
\bibliography{citation}
}




\end{document}